\documentclass{article}
\usepackage[preprint]{neurips_2026}

\usepackage[utf8]{inputenc}
\usepackage[T1]{fontenc}
\usepackage{hyperref}
\usepackage{url}
\usepackage{booktabs}
\usepackage{amsfonts}
\usepackage{amsmath}
\usepackage{mathtools}
\usepackage{amsthm}
\usepackage{nicefrac}
\usepackage{microtype}
\usepackage{xcolor}
\usepackage{graphicx}
\usepackage{algorithm}
\usepackage{algorithmic}
\usepackage{multirow}
\usepackage{multicol}
\usepackage{enumitem}

\theoremstyle{plain}

\theoremstyle{definition}

\theoremstyle{remark}

\title{DecompressionLM: Deterministic, Diagnostic, and\\Zero-Shot Concept Graph Extraction\\from Language Models}

\author{
  Zhaochen Hong\\
  Siebel School of Computing and Data Science\\
  University of Illinois Urbana-Champaign\\
  Urbana, IL, USA\\
  \texttt{zhong42@illinois.edu}\\
  \And
  Jiaxuan You\\
  Siebel School of Computing and Data Science\\
  University of Illinois Urbana-Champaign\\
  Urbana, IL, USA\\
  \texttt{jiaxuan@illinois.edu}\\
}

\begin{document}
\setlength{\textfloatsep}{7pt}
\setlength{\floatsep}{6pt}
\setlength{\intextsep}{6pt}
\renewcommand{\arraystretch}{0.93}

\maketitle

\begin{abstract}
Existing knowledge probing methods rely on predefined queries, which constrain extraction to known concepts and introduce context-dependent biases. We introduce DecompressionLM, a stateless framework for zero-shot concept graph extraction from large language models using deterministic sampling via Van der Corput sequences combined with arithmetic decoding. Rather than attempting to recover the full extent of parametric knowledge encoded in model weights, our approach characterizes the set of concepts that a model can reliably \emph{surface} under controlled, minimal prompting. We define \emph{concept coverage} as a diagnostic measure of the accessible support of the model's output distribution under this sampling protocol. Across two model families and multiple quantization variants, we find that activation-aware quantization (AWQ-4bit) consistently increases accessible concept diversity, while uniform quantization (GPTQ-Int4) leads to severe degradation and fragmentation, effects not reflected by perplexity. Normalized coverage metrics confirm that these differences persist after controlling for token budget and newline rate. Comparisons against i.i.d.\ arithmetic and ancestral sampling baselines show that VdC sampling achieves competitive concept coverage while consistently improving run-to-run reproducibility, validating its role as a structured exploration schedule rather than a coverage-inflating mechanism. Corpus-based verification further shows that hallucination rates correlate with MMLU-Pro Law rankings (Spearman $\rho=0.57$, $p<0.01$, 95\% bootstrap CI $[0.12, 0.81]$). Concept coverage provides a complementary diagnostic signal for model compression evaluation. Code is available at \url{https://github.com/ulab-uiuc/decompressionlm}.
\end{abstract}

\section{Introduction}

\begin{figure}[ht]
\begin{center}
\includegraphics[width=\textwidth]{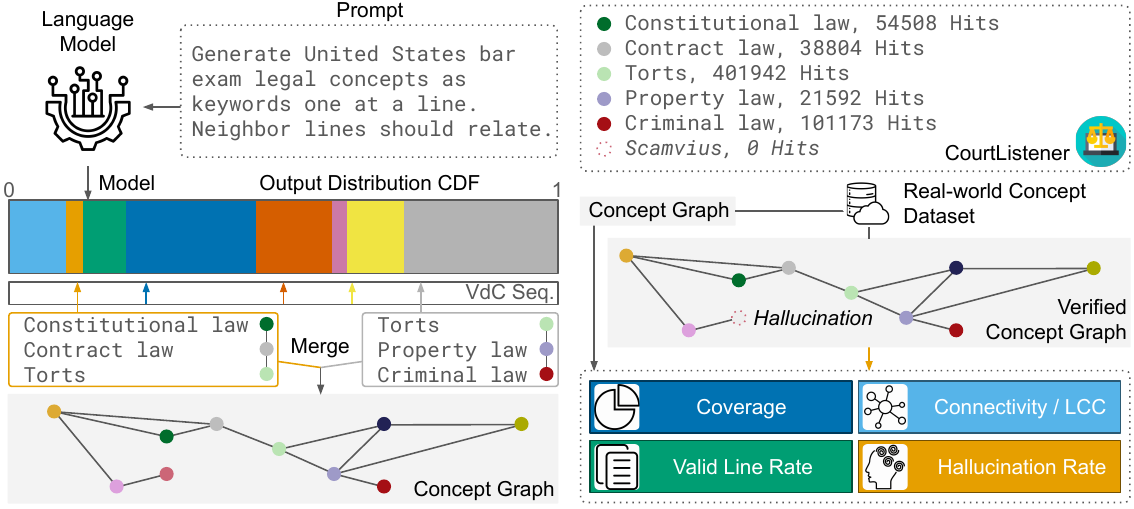}
\vspace{-16pt}
\caption{DecompressionLM pipeline. The Language Model is given a prompt asking for related concepts under a domain, while a VdC-based arithmetic decoding algorithm explores the distribution space and samples sequences faithfully; the sequences are parsed into chains of concepts and merged into a concept graph (left) and grounded against CourtListener for hallucination analysis (right).}
\label{fig:pipeline}
\end{center}
\end{figure}

Large language models (LLMs) encode substantial factual knowledge in their parameters \citep{brown2020,petroni-etal-2019-language}, yet measuring what a model can reliably \emph{surface} under controlled decoding protocols at scale remains challenging. Query-based probing \citep{petroni-etal-2019-language} requires pre-defined templates, bounding extraction to concepts researchers already know to ask about. Prompt engineering leaks information by naming target concepts. Sequential beam search introduces cross-hypothesis competition, rich-get-richer dynamics, and quadratic context-scaling costs, all of which bias exploration of accessible concepts \citep{holtzman2020,vaswani2023}.

We introduce \textbf{DecompressionLM}, a stateless, zero-shot framework for concept graph extraction that characterizes the \emph{accessible support} of a model's output distribution under a fixed, minimal prompting protocol. Rather than querying with pre-specified questions, we systematically sample the output space using Van der Corput (VdC) low-discrepancy codes \citep{niederreiter1992,dick2010} combined with arithmetic decoding \citep{vilnis2023arithmeticsamplingparalleldiverse}, enabling parallel, seed-free, deterministic generation without cross-sequence interference or shared decoding state.

Our key contributions are:

\begin{enumerate}[leftmargin=1.5em,itemsep=2pt]
\item \textbf{A reproducible, stateless probing protocol.} VdC sampling achieves competitive concept coverage while consistently improving run-to-run Jaccard stability ($0.316\pm0.040$ vs $0.279\pm0.023$ for i.i.d.\ arithmetic and $0.271\pm0.021$ for ancestral at $\ell=16$), validating its role as a structured exploration schedule rather than a coverage-inflating mechanism under parallel decoding.

\item \textbf{Stateless concept extraction with graph diagnostics.} Sequences are generated independently without shared context or cross-sequence competition. The concept graph captures concept diversity, connectivity, valid-line rate, and fragmentation patterns across quantization variants.

\item \textbf{Concept coverage as a complementary diagnostic signal.} Concept coverage exposes quantization effects invisible to perplexity. AWQ-4bit increases surfaced concept diversity, while GPTQ-Int4 induces two qualitatively distinct failure modes: semantic-only collapse in Qwen2.5-7B, and simultaneous semantic and structural collapse in Llama-3.1-8B under identical prompting and decoding conditions across matched sampling budgets and normalization procedures.\end{enumerate}

\section{Related Work}

\paragraph{Knowledge probing.} \citet{petroni-etal-2019-language} introduced LAMA, treating language models as knowledge bases via cloze-style prompts. AutoPrompt \citep{autoprompt} uses gradient-based search to discover optimal templates, while subsequent work has studied context effects \citep{petroni2020context} and negated probes \citep{kassner2020negated}. All these approaches share a closed-world limitation: extraction is bounded by pre-specified queries. Recent work \citep{noroozizadeh2024geometric} demonstrates that parametric memory organizes knowledge geometrically, motivating graph-structured extraction.

\paragraph{Decoding and sampling diversity.} \citet{holtzman2020} identify neural text degeneration in likelihood-maximizing decoding. Ancestral sampling \citep{giulianelli-etal-2023-information,amini2024structuredvoronoisampling,zhang-etal-2023-mixce} draws tokens directly from the model's distribution. Arithmetic Sampling \citep{arisam} uses CDF-based codes to guide sequence generation. Importance Sampling \citep{bai2017tapastwopassapproximateadaptive,anshumann-etal-2025-sparse}, Rejection Sampling \citep{ye2025efficientasymptoticallyunbiasedconstrained,pohl2025decodingfreesamplingstrategiesllm}, and MCMC \citep{gonzalez2025constrainedsamplinglanguagemodels,faria2025sampledontsearchrethinking} correct biased proposals via reweighting, accept-reject filtering, or Markov transitions. The choice of decoding strategy profoundly affects generation quality, diversity, coverage, stability, and long-tail concept accessibility patterns \citep{meister2023decoding}.

\paragraph{Model quantization.} GPTQ \citep{gptq} uses second-order information for optimal bit allocation; SmoothQuant \citep{smoothquant} migrates quantization difficulty from activations to weights. AWQ \citep{lin2024} protects salient weights based on activation distributions, achieving better accuracy preservation than uniform quantization at the same bit-width. Quantization-aware training \citep{onebit} and fully quantized training \citep{loqt} offer alternative compression strategies. Existing evaluations rely primarily on perplexity and downstream benchmarks; we introduce concept coverage as a complementary diagnostic evaluation dimension.

\section{Method}
\label{sec:method}

Our objective is not to optimize generation quality, but to define a controlled and reproducible sampling procedure for probing the support of the model's output distribution under fixed prompting and decoding conditions. Each component of the method is designed to minimize biases introduced by decoding strategies, cross-sequence interactions, and prompt conditioning.

\subsection{Problem Formulation}

Given a language model $M$ with vocabulary $\mathcal{V}$ and a domain description $d$, our goal is to extract a concept graph $G = (C, E)$ where $C$ is a set of valid concepts and $E$ represents relationships discovered through the model's knowledge structure. For a prompt template $p(d)$ conditioning the model on domain $d$, we seek:
\begin{equation}
C = \{c \mid P_M(c \mid p(d)) > \epsilon,\ \text{valid}(c)\},
\qquad
E = \{(c_i, c_j) \mid c_j \text{ discovered when exploring } c_i\},
\end{equation}
where $\text{valid}(\cdot)$ enforces syntactic constraints and the directed edge $(c_i, c_j)$ indicates co-occurrence in the same generated sequence.

Concept coverage measures accessible outputs under the specified prompting and decoding protocol rather than complete latent parametric knowledge.

\subsection{Motivation: Suppressing Context Leakage}
\label{sec:motivation}

DecompressionLM minimizes context leakage by using minimal domain prompts without naming target concepts, short independent sequences, and stateless decoding without cross-sequence interaction. This reduces contamination from prompt-specific hints, accumulated reasoning context, and beam-style competition effects \citep{liang2025the,min-etal-2022-rethinking,NEURIPS2022_9d560961}.

\subsection{Van der Corput Sampling with Arithmetic Decoding}
\label{sec:vdc-sampling}

\paragraph{Arithmetic decoding.} Arithmetic coding \citep{vilnis2023arithmeticsamplingparalleldiverse} provides a bijection between sequences and codes $z \in [0,1)$. For sequence $\mathbf{s} = (s_1, \ldots, s_T)$, decoding maintains an interval $[L, U)$ subdivided by token probabilities:
\begin{align}
L_0 &= 0,\quad U_0 = 1, \\
L_{t+1} &= L_t + (U_t - L_t)\cdot\mathrm{CDF}(s_{t+1} \mid \mathbf{s}_{<t}), \\
U_{t+1} &= L_t + (U_t - L_t)\cdot\mathrm{CDF}(s_{t+1}{+}1 \mid \mathbf{s}_{<t}),
\end{align}
where $\mathrm{CDF}(k \mid \mathbf{s}_{<t}) = \sum_{i=1}^{k-1} P_M(v_i \mid p(d), \mathbf{s}_{<t})$.

\paragraph{Van der Corput sequence.} Rather than drawing codes randomly, we use the Van der Corput (VdC) sequence in base 2:
\begin{equation}
v_n = \sum_{i=0}^{\lfloor \log_2 n \rfloor} \frac{d_i}{2^{i+1}},
\end{equation}
where $d_i$ are the binary digits of $n$. The star discrepancy satisfies $D_T^* = O(\log T / T)$ versus $O(T^{-1/2})$ for i.i.d.\ sampling \citep{niederreiter1992}, ensuring more uniform coverage of the arithmetic code space and empirically yielding more systematic exploration of generated outputs. Crucially, the VdC sequence is seed-free and prefix-consistent: the first $N$ points of the sequence are a fixed deterministic schedule regardless of total budget, enabling consistent comparisons across models, quantization variants, and runs without requiring a shared random seed. Empirically, we find that VdC achieves competitive concept coverage relative to i.i.d.\ arithmetic and ancestral sampling baselines while consistently improving run-to-run reproducibility as measured by inter-run Jaccard similarity (Section~\ref{sec:exp4-baselines}) across multiple sequence lengths and quantization configurations in our experiments.

For robustness evaluation, we apply a phase shift $v_n^{(\phi)} = (v_n + \phi) \bmod 1$ and report metrics averaged across multiple offsets $\phi$.

\paragraph{Parallel stateless generation.} Each code $v_n$ defines an independent sequence generation from prompt $p(d)$. All $N$ sequences $\{\mathbf{s}_n\}_{n=1}^N$ are produced independently without coordination, enabling embarrassingly parallel execution. Full implementation details including EOS handling, tokenizer settings, and determinism conditions are provided in Appendix~\ref{sec:arithmetic-impl}.

\subsection{Concept Extraction and Validation}
\label{sec:concept-validation}

Decoded sequences are parsed into ordered concept candidates (one per line), then processed by the multi-stage pipeline in Algorithm~\ref{alg:concept-extraction}.

\begin{algorithm}[t]
\caption{Concept Extraction and Graph Construction}
\label{alg:concept-extraction}
\begin{algorithmic}
\STATE {\bfseries Input:} Sequences $\{\mathbf{s}_n\}_{n=1}^N$, similarity threshold $\tau=90$
\STATE {\bfseries Output:} Directed graph $G=(V,E)$
\STATE \textcolor{gray}{// Phase 1: Extract and normalize}
\FOR{each sequence $\mathbf{s}_n$}
  \STATE Decode to text, split on newlines $\to \{c_1, \ldots, c_m\}$; discard truncated final concept
  \STATE Normalize each $c_i$: NFKC, lowercase, singularize head noun
  \STATE Add edges $\{(c_i, c_{i+1})\}$ to $E_{\text{raw}}$
\ENDFOR
\STATE \textcolor{gray}{// Phase 2: Fuzzy merge with Levenshtein clustering ($\tau=90$)}
\STATE $\phi \gets \textsc{FuzzyMerge}(C_{\text{raw}}, \tau)$
\STATE \textcolor{gray}{// Phase 3: Build merged graph}
\FOR{each edge $(c_i, c_j) \in E_{\text{raw}}$}
  \STATE $w(\phi(c_i), \phi(c_j)) \mathrel{+}= 1$ \; if $\phi(c_i) \neq \phi(c_j)$
\ENDFOR
\STATE \textbf{return} $G$
\end{algorithmic}
\end{algorithm}

\paragraph{Normalization and merging.} Each raw concept undergoes Unicode normalization (NFKC), lowercasing, ASCII filtering, and morphological singularization of the head noun using the \texttt{inflect} library \citep{inflect2024}. Near-duplicates are merged via Levenshtein-based clustering with length-based blocking, using similarity:
\begin{equation}
\mathrm{sim}(c_i, c_j) = 100 \times \left(1 - \frac{\mathrm{Lev}(c_i, c_j)}{\max(|c_i|, |c_j|)}\right).
\end{equation}
Pairs with $\mathrm{sim} \geq 90$ are merged to a canonical representative.

\paragraph{Corpus-grounded validation.} For the US Law domain, we verify extracted concepts against CourtListener \citep{CourtListener}, which indexes the majority of published U.S.\ case law. A concept $c \in V$ is verified if it appears in at least one legal document; otherwise it is classified as a hallucination.

\section{Experiments}

\subsection{Setup}

\paragraph{Models and quantization.} We evaluate \textbf{Qwen2.5-7B-Instruct} \citep{qwen2.5} and \textbf{Llama-3.1-8B-Instruct} \citep{llama3} under five configurations: \textbf{BF16} (baseline), \textbf{GPTQ-Int8} \citep{gptq}, \textbf{GPTQ-Int4} \citep{gptq}, \textbf{AWQ-4bit} \citep{lin2024}, and \textbf{BNB-4bit} \citep{qlora}.

\paragraph{Domains and sampling.} We extract concepts from four domains: US bar examination law, Git/version control, radiology and medical imaging, and FAA aviation regulations. Each run uses $N=8192$ sequences with maximum length $\ell \in \{16, 32\}$ tokens, generated on a single NVIDIA A100 (80GB).

\paragraph{Hallucination evaluation.} We evaluate 21 open-source instruction-tuned models ($\leq$16B parameters) ranked by MMLU-Pro Law performance \citep{mmlupro}. For each model we generate $N=8192$ sequences and uniformly sample 200 concepts for CourtListener verification.

We emphasize that all evaluations are conducted under a fixed sampling protocol; results should be interpreted as properties of accessible outputs under this protocol, rather than intrinsic properties of the underlying model.

\subsection{Experiment 1: Quantization Effects on Concept Coverage}

\begin{table}[t]
\caption{Concept graph statistics for Qwen2.5-7B-Instruct and Llama-3.1-8B-Instruct on US Law. AWQ-4bit extracts 2.7$\times$ more concepts than BF16 for Qwen while maintaining connectivity; GPTQ-Int4 exhibits severe collapse. \textbf{Nodes}: unique concepts; \textbf{Edges}: co-occurrence links; \textbf{Deg}: mean degree; \textbf{Comp}: connected components; \textbf{CC\%}: largest component fraction.}
\label{tab:quant-comparison}
\centering
\small
\begin{tabular}{lrrrrrr}
\toprule
\multicolumn{7}{c}{\textbf{Qwen2.5-7B-Instruct} ($\ell=16$)} \\
\cmidrule(r){1-7}
Variant & Nodes & Edges & Density & Deg & Comp & CC\% \\
\midrule
AWQ-4bit  & 1,052 & 2,736 & 0.0025 & 5.20 &  12 & 96.5 \\
BF16      &   384 &   988 & 0.0067 & 5.15 &   8 & 95.3 \\
GPTQ-Int8 &   371 &   915 & 0.0067 & 4.93 &   8 & 94.6 \\
BNB-4bit  &   335 &   808 & 0.0072 & 4.82 &   8 & 93.7 \\
GPTQ-Int4 &    55 &    85 & 0.0286 & 3.09 &   3 & 87.3 \\
\midrule
\multicolumn{7}{c}{\textbf{Qwen2.5-7B-Instruct} ($\ell=32$)} \\
\cmidrule(r){1-7}
AWQ-4bit  & 1,899 & 7,408 & 0.0021 & 7.80 &   4 & 99.7 \\
BF16      &   809 & 3,273 & 0.0050 & 8.09 &   6 & 97.4 \\
GPTQ-Int8 &   786 & 3,132 & 0.0051 & 7.97 &   3 & 98.9 \\
BNB-4bit  &   680 & 2,408 & 0.0052 & 7.08 &   5 & 98.5 \\
GPTQ-Int4 &   101 &   211 & 0.0209 & 4.18 &   2 & 97.0 \\
\midrule
\multicolumn{7}{c}{\textbf{Llama-3.1-8B-Instruct} ($\ell=16$)} \\
\cmidrule(r){1-7}
AWQ-4bit  &  6,878 & 11,555 & 0.00024 & 3.36 &   601 & 80.9 \\
BF16      &  5,259 & 10,841 & 0.00039 & 4.12 &   327 & 86.6 \\
GPTQ-Int8 &  5,241 & 10,771 & 0.00039 & 4.11 &   349 & 85.8 \\
BNB-4bit  &  4,029 &  7,524 & 0.00046 & 3.73 &   275 & 85.3 \\
GPTQ-Int4 &  1,502 &  1,215 & 0.00054 & 1.62 &   391 & 40.8 \\
\midrule
\multicolumn{7}{c}{\textbf{Llama-3.1-8B-Instruct} ($\ell=32$)} \\
\cmidrule(r){1-7}
AWQ-4bit  & 14,654 & 34,439 & 0.00016 & 4.70 &   147 & 96.7 \\
BF16      & 10,922 & 32,220 & 0.00027 & 5.90 &    78 & 97.6 \\
GPTQ-Int8 & 10,991 & 31,949 & 0.00026 & 5.81 &    73 & 97.9 \\
BNB-4bit  &  9,167 & 22,821 & 0.00027 & 4.98 &   103 & 96.4 \\
GPTQ-Int4 &  5,360 &  4,730 & 0.00017 & 1.76 & 1,097 & 51.1 \\
\bottomrule
\end{tabular}
\end{table}

Results in Table~\ref{tab:quant-comparison} reveal a striking divergence. AWQ-4bit extracts 2.7$\times$ more concepts than BF16 for Qwen at $\ell=16$ while maintaining high connectivity (96.5\% largest component). For Llama, AWQ-4bit expands coverage by +30.8\%. In contrast, GPTQ-Int4 exhibits severe fragmentation: only 40.8\% of Llama concepts reside in the largest component at $\ell=16$, indicating disruption of graph connectivity structure rather than mere concept reduction. GPTQ-Int8 performs near-identically to BF16 throughout.

\paragraph{Cross-domain consistency.} AWQ's advantage generalizes across all four domains: $2.74{\times}$ BF16 for US Law, $1.43{\times}$ for Git, $2.12{\times}$ for Radiology, and $1.56{\times}$ for FAA. This consistency across structurally distinct domains suggests the effect reflects how quantization methods preserve general access to diverse knowledge. AWQ's superior performance aligns with its design: by protecting only the 1\% of weights with highest activation magnitudes \citep{lin2024}, it appears to preserve or redistribute probability mass in a way that increases the accessibility of diverse, long-tail concepts in the output distribution.

\paragraph{Perplexity decoupling.} Table~\ref{tab:perplexity-results} shows that mean/median explanation-level perplexity remains within a narrow range across quantization variants in each domain, even when concept coverage diverges dramatically. This partial decoupling suggests that concept accessibility (whether a concept is surfaced under open-ended sampling) and local explanation fluency (once the concept is provided in the prompt) are distinct properties, and that perplexity alone does not fully capture changes in accessible knowledge induced by compression.

\begin{table}[t]
\caption{Perplexity statistics for Llama-3.1-8B-Instruct when explaining extracted concepts ($\ell=16$, $N=200$ per variant). Mean/median perplexity remains stable across variants despite large coverage differences. GPTQ-Int4 perplexities are shown at order-of-magnitude level to reflect a degenerate regime.}
\label{tab:perplexity-results}
\centering
\small
\begin{tabular}{lrrrr}
\toprule
Variant & Mean PPL & Med PPL & Std & FreqCorr \\
\midrule
\multicolumn{5}{c}{\textit{US Law}} \\
BF16       & 2.51 & 2.44 & 0.44 & $-0.17$ \\
GPTQ-Int8  & 2.55 & 2.46 & 0.51 & $-0.24$ \\
AWQ-4bit   & 2.67 & 2.65 & 0.49 & $-0.24$ \\
BNB-4bit   & 2.53 & 2.48 & 0.55 & $-0.24$ \\
GPTQ-Int4  & $\sim\!10^{5}$ & $\sim\!10^{4}$ & $\sim\!10^{5}$ & $-0.07$ \\
\midrule
\multicolumn{5}{c}{\textit{Radiology Imaging}} \\
BF16       & 2.36 & 2.30 & 0.43 & $-0.32$ \\
GPTQ-Int8  & 2.37 & 2.27 & 0.59 & $-0.30$ \\
AWQ-4bit   & 2.46 & 2.38 & 0.48 & $-0.34$ \\
BNB-4bit   & 2.34 & 2.24 & 0.47 & $-0.34$ \\
GPTQ-Int4  & $\sim\!10^{5}$ & $\sim\!10^{4}$ & $\sim\!10^{5}$ & $-0.05$ \\
\bottomrule
\end{tabular}
\end{table}

\subsection{Experiment 2: Robustness to VdC Offset Variation}
\label{sec:exp2-vdc-offsets}

Across 8 VdC offsets, the quantization degradation pattern remains stable for both model families. Pairwise Jaccard similarities range from 15 to 34\%, and GPTQ-Int4 consistently yields the fewest core concepts, confirming that the collapse is not an artifact of a particular VdC offset. Full offset statistics are reported in Appendix~\ref{sec:appendix-vdc}.

\subsection{Experiment 3: Corpus-Grounded Hallucination Analysis}

\begin{table}[t]
\caption{Hallucination rates for 21 MMLU-Pro Law ranked models with 95\% Wilson confidence intervals. Better benchmark performance correlates with lower hallucination (Spearman $\rho=0.571$, 95\% CI $[0.122, 0.806]$, $p=0.007$; Pearson $r=0.525$, $p=0.014$). FreqCorr: within-model Pearson correlation between concept frequency and corpus verification.}
\label{tab:hallucination}
\centering
\small
\begin{tabular}{clrrrr}
\toprule
Rank & Model & Hall\% & 95\% CI & Verified & FreqCorr \\
\midrule
1 & Gemma-2-9B         & 12.1 & [8.2, 17.3]  & 175 & $+0.154$ \\
2 & Mistral-Nemo       & 19.3 & [14.4, 25.4] & 159 & $+0.204$ \\
3 & Phi-3.5-mini       & 14.5 & [10.3, 20.0] & 171 & $+0.178$ \\
4 & Qwen2-7B           & 12.1 & [8.2, 17.3]  & 175 & $+0.164$ \\
5 & Phi-3-mini-4k      & 18.5 & [13.7, 24.5] & 163 & $+0.229$ \\
\midrule
6 & Llama-3.1-8B       & 28.1 & [22.4, 34.8] & 143 & $+0.227$ \\
7 & Phi-3-mini-128k    & 13.5 & [9.4, 18.9]  & 173 & $+0.180$ \\
8 & Llama-3-8B         & 33.2 & [27.0, 40.0] & 133 & $-0.088$ \\
9 & Llama-3-Smaug      & 56.8 & [49.8, 63.5] &  86 & $-0.144$ \\
10 & Granite-3.1-8B    & 28.0 & [22.2, 34.6] & 144 & $+0.184$ \\
\midrule
11 & Ministral-8B      & 23.5 & [18.1, 29.9] & 150 & $+0.250$ \\
12 & EXAONE-3.5-2.4B   & 46.7 & [39.9, 53.7] & 106 & $+0.089$ \\
13 & Granite-3.1-2B    & 32.8 & [26.6, 39.7] & 131 & $+0.086$ \\
14 & Mistral-7B        & 27.5 & [18.9, 38.1] &  58 & $+0.273$ \\
15 & DeepSeek-V2-Lite  & 32.8 & [26.7, 39.6] & 133 & $+0.136$ \\
\midrule
16 & Neo-7B            & 22.2 & [17.0, 28.5] & 154 & $+0.149$ \\
17 & Granite-3.1-3B    & 44.7 & [37.9, 51.6] & 109 & $+0.153$ \\
18 & Qwen2-1.5B        & 27.8 & [22.0, 34.4] & 143 & $+0.186$ \\
19 & Qwen2-0.5B        & 34.2 & [27.8, 41.2] & 125 & $+0.198$ \\
20 & DeepSeek-Math-7B  & 25.5 & [20.0, 32.0] & 149 & $+0.279$ \\
21 & Granite-3.1-1B    & 42.4 & [35.7, 49.4] & 114 & $+0.294$ \\
\bottomrule
\end{tabular}
\end{table}

Across all 21 models, the average hallucination rate is 27.8\%, meaning 72.2\% of sampled concepts are corpus-verified, indicating that a substantial fraction of surfaced concepts correspond to corpus-grounded legal concepts. Table~\ref{tab:hallucination} shows systematic variation aligned with MMLU-Pro Law rankings. Top-5 models average 15.3\% hallucination versus 34.9\% for the bottom-5, a 19-point gap ($\chi^2=100.4$, $p{<}0.001$). Spearman rank correlation across all 21 models is $\rho=0.571$ (95\% bootstrap CI $[0.122, 0.806]$, $p=0.007$), and Pearson $r=0.525$ (95\% CI $[0.172, 0.820]$, $p=0.014$), confirming a statistically significant relationship between benchmark rank and hallucination rate. Per-model 95\% Wilson confidence intervals are reported in Table~\ref{tab:hallucination}.

Two qualitatively distinct failure modes emerge. \emph{Tail collapse} (Granite-3.1-3B at 44.7\%, EXAONE-3.5-2.4B at 46.7\%, both with positive FreqCorr): rare concepts are disproportionately hallucinated while common concepts remain grounded. \emph{Head corruption} (Llama-3-Smaug at 56.8\%, FreqCorr$=-0.144$; Llama-3-8B at 33.2\%, FreqCorr$=-0.088$): even frequently generated concepts become unreliable, suggesting fundamental disruption of the frequency-knowledge relationship rather than simple knowledge sparsity.

\subsection{Experiment 4: Sampling Baseline Comparison}
\label{sec:exp4-baselines}

A natural question is whether the observed coverage properties are specific to VdC-arithmetic decoding or arise under any sampling schedule. We compare three methods on Qwen2.5-7B-Instruct BF16 over US Law at $\ell \in \{16, 32\}$, with $N=2048$ sequences per run and 8 independent runs each:

\begin{itemize}[leftmargin=1.5em,itemsep=2pt]
\item \textbf{VdC-arithmetic}: Van der Corput codes with arithmetic decoding; 8 runs use distinct Cranley-Patterson offsets $\phi$.
\item \textbf{Random-arithmetic}: i.i.d.\ $\mathcal{U}(0,1)$ codes with the same arithmetic decoder.
\item \textbf{Ancestral}: \texttt{model.generate} with $T{=}1$, top-$p{=}1$, top-$k{=}0$.
\end{itemize}

We report unique concepts (mean $\pm$ std across 8 runs), coverage normalised by total tokens generated (Cov/10kT) and by total parsed lines (Cov/1kL), valid line rate (VLR), and pairwise Jaccard similarity across runs.

\begin{table}[t]
\caption{Sampling method comparison on Qwen2.5-7B-Instruct BF16, US Law ($N=2048$ per run, 8 runs). All metrics are mean\,$\pm$\,std across 8 independent runs.
\textbf{VdC}: Van der Corput arithmetic (paper method);
\textbf{Rand}: i.i.d.\ uniform arithmetic codes;
\textbf{Anc}: ancestral sampling ($T{=}1$, top-$p{=}1$).
Cov/10kT: concepts per 10k tokens generated; Cov/1kL: concepts per 1k parsed lines; VLR: valid line rate; Jacc: pairwise Jaccard across runs.}
\label{tab:sampling-baselines}
\centering\small
\begin{tabular}{llrrrrr}
\toprule
$\ell$ & Method & Concepts & Cov/10kT & Cov/1kL & VLR & Jacc \\
\midrule
16 & VdC  & $164.4\pm9.2$  & $126.18\pm6.53$  & $36.10\pm1.94$ & $0.975\pm0.002$ & $0.316\pm0.040$ \\
16 & Rand & $161.6\pm13.0$ & $123.05\pm10.12$ & $35.32\pm3.00$ & $0.976\pm0.004$ & $0.279\pm0.023$ \\
16 & Anc  & $154.8\pm9.5$  & $118.98\pm7.24$  & $34.25\pm2.09$ & $0.976\pm0.002$ & $0.271\pm0.021$ \\
\midrule
32 & VdC  & $356.6\pm24.1$ & $159.96\pm10.25$ & $50.70\pm3.29$ & $0.966\pm0.002$ & $0.293\pm0.025$ \\
32 & Rand & $358.2\pm16.7$ & $159.43\pm8.10$  & $50.65\pm2.51$ & $0.966\pm0.003$ & $0.281\pm0.014$ \\
32 & Anc  & $365.9\pm16.0$ & $162.06\pm6.40$  & $52.04\pm2.05$ & $0.970\pm0.003$ & $0.275\pm0.020$ \\
\bottomrule
\end{tabular}
\end{table}

Results in Table~\ref{tab:sampling-baselines} show that VdC-arithmetic achieves concept coverage comparable to both random-arithmetic and ancestral sampling while consistently producing higher cross-run Jaccard similarity ($0.316\pm0.040$ vs $0.279\pm0.023$ and $0.271\pm0.021$ at $\ell=16$). VLR remains stable across all three methods ($\approx0.975$), confirming that formatting differences do not explain the observed trends. These results support VdC's role as a structured, reproducible exploration schedule rather than a coverage-inflating mechanism.

\subsection{Experiment 5: Coverage Normalised by Tokens and Lines}
\label{sec:exp5-normalized}

Raw unique-concept counts may be confounded by differences in newline generation rate or per-sequence token budget across quantization variants. To address this, we re-read the existing Experiment~1 parquet files (no model inference required) and compute three normalised coverage metrics designed to isolate semantic diversity from formatting and tokenization effects across variants:

\begin{align}
\text{Cov}/10\text{kT} &= \frac{\text{unique concepts}}{\text{total tokens}} \times 10000, \\
\text{Cov}/1\text{kL}  &= \frac{\text{unique concepts}}{\text{total parsed lines}} \times 1000, \\
\text{VLR}             &= \frac{\text{lines surviving normalisation}}{\text{total parsed lines}}.
\end{align}

These metrics control for the possibility that AWQ-4bit simply produces more newlines or shorter sequences, artificially inflating the raw concept count through formatting-related effects alone.

\begin{table}[t]
\caption{Normalised coverage for Qwen2.5-7B-Instruct and Llama-3.1-8B-Instruct on US Law at $\ell=16$. Cov/10kT and Cov/1kL control for differences in token budget and newline rate. VLR: fraction of parsed lines yielding a valid concept after normalisation. GPTQ-Int4 Llama VLR of 0.527 indicates simultaneous formatting and semantic collapse.}
\label{tab:coverage-normalized}
\centering\small
\begin{tabular}{llrrrrr}
\toprule
Model & Variant & Concepts & Cov/10kT & Cov/1kL & VLR & LCC\% \\
\midrule
\multirow{5}{*}{Qwen2.5-7B}
& AWQ-4bit   & 1,052 & 80.94 & 25.44 & 0.965 & 96.5 \\
& BF16       &   384 & 73.05 & 20.95 & 0.976 & 95.3 \\
& GPTQ-Int8  &   371 & 75.85 & 21.41 & 0.976 & 94.6 \\
& BNB-4bit   &   335 & 80.24 & 22.64 & 0.976 & 93.7 \\
& GPTQ-Int4  &    55 & 28.73 &  6.51 & 0.992 & 87.3 \\
\midrule
\multirow{5}{*}{Llama-3.1-8B}
& AWQ-4bit   & 6,878 & 527.74 & 240.24 & 0.984 & 80.9 \\
& BF16       & 5,259 & 411.94 & 188.25 & 0.988 & 86.6 \\
& GPTQ-Int8  & 5,241 & 411.41 & 187.79 & 0.986 & 85.8 \\
& BNB-4bit   & 4,029 & 391.65 & 181.68 & 0.987 & 85.3 \\
& GPTQ-Int4  & 1,502 & 114.59 &  82.96 & 0.527 & 40.8 \\
\bottomrule
\end{tabular}
\end{table}

\begin{figure}[t]
\centering
\includegraphics[width=\linewidth]{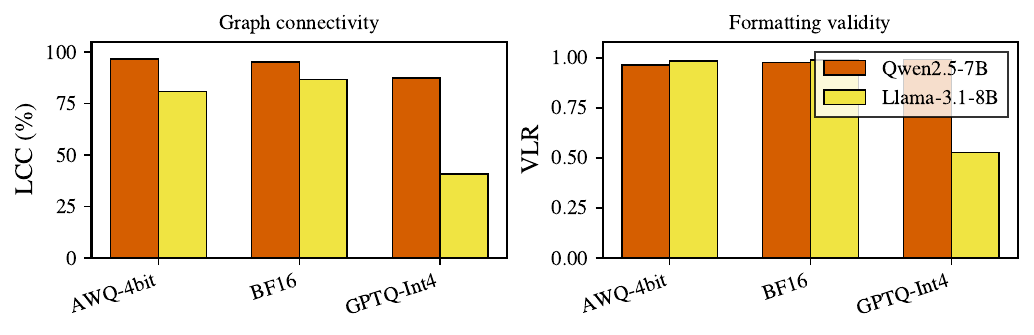}
\vspace{-24pt}
\caption{Structural and formatting degradation under quantization, US Law ($\ell=16$). GPTQ-Int4 preserves VLR for Qwen2.5-7B despite severe coverage loss, but collapses both VLR and graph connectivity for Llama-3.1-8B.}
\label{fig:structural-collapse}
\end{figure}

Table~\ref{tab:coverage-normalized} and Figure~\ref{fig:structural-collapse} show that AWQ's advantage persists after normalization; the increase in surfaced concept diversity is not solely explained by formatting or token-budget effects. GPTQ-Int4 exhibits distinct degradation patterns: Qwen2.5-7B shows semantic collapse despite near-normal VLR (0.992), while Llama-3.1-8B exhibits simultaneous semantic, formatting, and graph-connectivity collapse (VLR$=0.527$, LCC$=40.8\%$). These degradation modes are not reflected by perplexity alone.

\subsection{Experiment 6: Prompt Paraphrase and Ranking Robustness}
\label{sec:prompt-robustness}

We evaluate whether coverage measurements and quantization orderings are sensitive to prompt surface form using five semantically equivalent paraphrases (all requesting legal \emph{concepts}, differing only in phrasing) on Qwen2.5-7B-Instruct BF16 ($\ell=16$, $N=2048$, 8 runs each), and three of these prompts across AWQ-4bit, BF16, and GPTQ-Int4 for both model families under identical decoding and normalization conditions across matched sampling budgets. More details are in Appendix~\ref{sec:appendix-prompt-paraphrase}.

Within-prompt Jaccard stability is consistent across variants ($0.266$--$0.328$), VLR remains high ($0.975$--$0.983$), and the mean cross-prompt Jaccard is $0.219 \pm 0.022$. Different paraphrases expose partially distinct regions of accessible concept space while sharing a common domain signal, consistent with our framing of concept coverage as a property of the model/prompt/decoder triad. Table~\ref{tab:prompt-ranking} shows that the AWQ-4bit $>$ BF16 $>$ GPTQ-Int4 ordering holds across all three prompts and both model families, confirming that the primary quantization findings are not artifacts of a single prompt wording.

\begin{table}[t]
\caption{Cross-variant prompt ranking robustness. AWQ-4bit $>$ BF16 $>$ GPTQ-Int4 holds for all 6/6 prompt-model combinations. Entries are mean concept counts across 4 runs.}
\label{tab:prompt-ranking}
\centering\small
\begin{tabular}{llrrrr}
\toprule
Prompt & Model & AWQ-4bit & BF16 & GPTQ-Int4 & OK \\
\midrule
Baseline      & Llama-3.1-8B & 2,318.5 & 1,865.5 & 418.5 & \checkmark \\
Baseline      & Qwen2.5-7B   &   424.5 &   160.8 &  15.5 & \checkmark \\
List relevant & Llama-3.1-8B & 2,014.0 & 1,648.2 & 325.8 & \checkmark \\
List relevant & Qwen2.5-7B   &   607.0 &   168.8 &  18.5 & \checkmark \\
Concise       & Llama-3.1-8B & 2,231.5 & 1,854.5 & 337.8 & \checkmark \\
Concise       & Qwen2.5-7B   &   552.0 &   131.2 &  42.2 & \checkmark \\
\bottomrule
\end{tabular}
\end{table}

Additional merge-threshold ($\tau \in \{85,90,95\}$) and top-$p$ ablations are reported in Appendix~\ref{sec:appendix-threshold}--\ref{sec:appendix-topp}; they preserve the main quantization ordering and show the expected coverage-stability tradeoff.

\subsection{Experiment 7: Coverage vs.\ Sampling Budget}
\label{sec:exp7-saturation}

We evaluate concept coverage as a function of sampling budget $N$ by recomputing concept counts from the first $N$ sequences of the Experiment~1 parquets for $N \in \{256, 512, 1024, 2048, 4096, 8192\}$.

\begin{figure}[t]
\centering
\includegraphics[width=\linewidth]{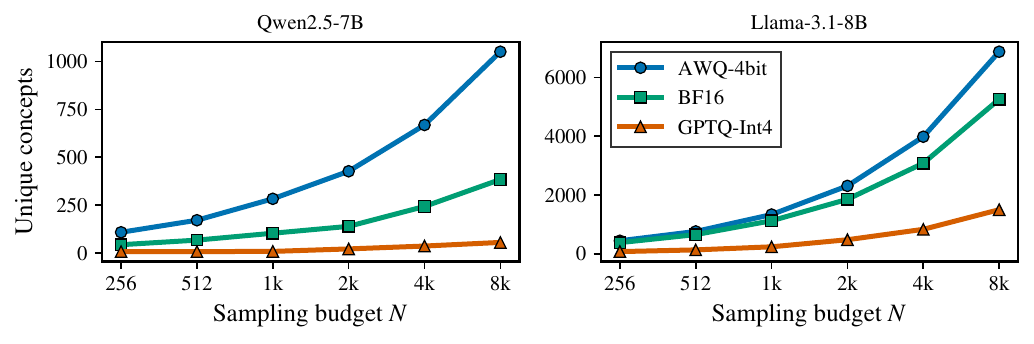}
\vspace{-24pt}
\caption{Concept coverage vs.\ sampling budget $N$, US Law ($\ell=16$). Ordering holds from $N=256$ to $N=8192$ (12/12 checks); growth decelerates at large $N$.}
\label{fig:saturation}
\end{figure}

Figure~\ref{fig:saturation} shows the ordering holds from $N=256$ onward (12/12 checks); marginal rates decrease across all variants (Qwen AWQ: 0.246$\to$0.094; Llama AWQ: 1.238$\to$0.706), confirming soft saturation rather than unbounded growth. Importantly, GPTQ-Int4 does not recover with larger $N$: even at $N=8192$ it reaches only 55 concepts for Qwen and 1,502 for Llama, well below BF16 at any tested budget. The gap between AWQ-4bit and BF16 also widens with $N$ for both model families, with Llama AWQ reaching 6,878 concepts versus BF16's 5,259 at full budget, suggesting that AWQ's long-tail diversity advantage accumulates rather than saturates over larger budgets. This finding is further reinforced by pairwise overlap analysis (Appendix~\ref{sec:appendix-overlap}): at $N=8192$, GPTQ-Int4 shares at most 5 concepts with any other Llama variant (Jaccard $\leq 0.1\%$), confirming that its accessible concept space is qualitatively distinct rather than merely smaller. Full counts in Appendix~\ref{sec:appendix-threshold}.

\section{Conclusion}

DecompressionLM characterizes accessible concept diversity via deterministic Van der Corput arithmetic decoding, exposing compression effects invisible to perplexity. Across two model families and four domains, VdC improves run-to-run stability relative to i.i.d.\ arithmetic and ancestral sampling at comparable coverage under controlled decoding conditions and fixed minimal prompting protocols. AWQ-4bit consistently increases surfaced concept diversity after normalization, while GPTQ-Int4 induces severe semantic and structural collapse not reflected by perplexity. Hallucination rates correlate with MMLU-Pro Law rankings (Spearman $\rho=0.571$, $p=0.007$), and 72.2\% of sampled concepts are corpus-verified. These results suggest that concept coverage provides a complementary diagnostic signal for evaluating quantization-induced changes in accessible model knowledge.

\paragraph{Limitations.}
Concept coverage is protocol-dependent and measures accessible support under a fixed model--prompt--decoder triad rather than complete latent parametric knowledge. Prompt paraphrases expose partially distinct concept sets ($J\approx0.22$), concept extraction and fuzzy normalization remain heuristic, and experiments focus primarily on 7B--8B quantization studies plus open models up to 16B in the hallucination analysis under fixed decoding budgets. CourtListener verification measures surface corpus occurrence rather than semantic correctness, and the mechanisms underlying GPTQ-Int4 failure modes remain an open question for future investigation across broader model scales.

\begin{ack}
Acknowledgments omitted for anonymous submission.
\end{ack}

\bibliographystyle{plainnat}
\bibliography{references}

\section*{Broader Impacts}

DecompressionLM is intended as a diagnostic framework for analyzing accessible semantic support under controlled decoding protocols. Potential positive impacts include improved reproducibility in model auditing, more informative evaluation of quantization methods beyond perplexity-based metrics, and detection of semantic or structural fragmentation under compression before deployment.

Coverage measurements are protocol-dependent and should not substitute expert validation in high-stakes settings. CourtListener verification addresses surface-level hallucination but not semantic correctness. We store only aggregate verification counts and do not redistribute CourtListener content.

\appendix

\section{Prompt Paraphrase Full Metrics}
\label{sec:appendix-prompt-paraphrase}

\begin{table}[h]
\caption{Full per-prompt paraphrase metrics on Qwen2.5-7B-Instruct BF16, US Law ($\ell=16$, $N=2048$, 8 runs). All variants request legal \emph{concepts}; only surface phrasing differs. Mean cross-prompt Jaccard: $0.219 \pm 0.022$.}
\label{tab:prompt-paraphrase-full}
\centering\small
\begin{tabular}{lrrrrr}
\toprule
Prompt & Concepts & Cov/10kT & VLR & Within-Jacc & LCC\% \\
\midrule
Baseline        & $164.4\pm9.2$  & $126.18\pm6.53$  & $0.975\pm0.002$ & $0.316\pm0.040$ & $95.2\pm3.8$ \\
List relevant   & $166.6\pm9.9$  & $176.32\pm9.97$  & $0.979\pm0.003$ & $0.328\pm0.027$ & $96.5\pm2.3$ \\
Study materials & $101.8\pm9.0$  & $144.57\pm11.78$ & $0.983\pm0.002$ & $0.266\pm0.043$ & $82.6\pm5.3$ \\
Concise         & $127.4\pm10.1$ & $139.23\pm10.06$ & $0.975\pm0.002$ & $0.307\pm0.033$ & $93.0\pm2.9$ \\
Study concepts  & $121.9\pm9.9$  & $145.15\pm11.24$ & $0.981\pm0.003$ & $0.289\pm0.035$ & $90.8\pm5.0$ \\
\bottomrule
\end{tabular}
\end{table}

\section{Merge Threshold Sensitivity and Saturation Counts}
\label{sec:appendix-threshold}

\begin{table}[h]
\caption{Concept coverage vs.\ sampling budget $N$, full counts. Same data as Figure~\ref{fig:saturation}.}
\label{tab:saturation}
\centering\small
\begin{tabular}{llrrrrrr}
\toprule
Model & Variant & $N{=}256$ & $N{=}512$ & $N{=}1024$ & $N{=}2048$ & $N{=}4096$ & $N{=}8192$ \\
\midrule
\multirow{3}{*}{Qwen2.5-7B}
& AWQ-4bit  &   108 &   171 &   283 &   427 &   669 & 1,052 \\
& BF16      &    43 &    67 &   103 &   139 &   243 &   384 \\
& GPTQ-Int4 &     7 &     7 &     8 &    21 &    36 &    55 \\
\midrule
\multirow{3}{*}{Llama-3.1-8B}
& AWQ-4bit  &   445 &   762 & 1,337 & 2,312 & 3,985 & 6,878 \\
& BF16      &   381 &   654 & 1,126 & 1,855 & 3,084 & 5,259 \\
& GPTQ-Int4 &    76 &   134 &   239 &   480 &   834 & 1,502 \\
\bottomrule
\end{tabular}
\end{table}

\begin{table}[h]
\caption{Concept node counts under merge thresholds $\tau \in \{85, 90, 95\}$ on Qwen2.5-7B and Llama-3.1-8B, US Law, $\ell=16$. The ordering AWQ-4bit $>$ BF16 $>$ GPTQ-Int4 is preserved across all values (12/12 checks); max change below 1\%.}
\label{tab:threshold-ablation}
\centering\small
\begin{tabular}{llrrr}
\toprule
Model & Variant & $\tau=85$ & $\tau=90$ & $\tau=95$ \\
\midrule
\multirow{5}{*}{Qwen2.5-7B}
& AWQ-4bit   & 1,049 & 1,052 & 1,052 \\
& BF16       &   384 &   384 &   384 \\
& GPTQ-Int8  &   370 &   371 &   371 \\
& BNB-4bit   &   334 &   335 &   335 \\
& GPTQ-Int4  &    55 &    55 &    55 \\
\midrule
\multirow{5}{*}{Llama-3.1-8B}
& AWQ-4bit   & 6,817 & 6,878 & 6,878 \\
& BF16       & 5,212 & 5,259 & 5,259 \\
& GPTQ-Int8  & 5,190 & 5,241 & 5,241 \\
& BNB-4bit   & 4,002 & 4,029 & 4,029 \\
& GPTQ-Int4  & 1,500 & 1,502 & 1,502 \\
\bottomrule
\end{tabular}
\end{table}

\section{Top-$p$ Nucleus Sampling Baselines}
\label{sec:appendix-topp}

\begin{table}[h]
\caption{Extended sampling comparison including top-$p$ nucleus sampling baselines on Qwen2.5-7B-Instruct BF16, US Law ($N=2048$, 8 runs). Top-$p$ truncation trades coverage for reproducibility; VdC sits at a distinct operating point with ancestral-level coverage and above-ancestral Jaccard.}
\label{tab:sampling-extended}
\centering\small
\begin{tabular}{llrrrrr}
\toprule
$\ell$ & Method & Concepts & Cov/10kT & VLR & Jacc & LCC\% \\
\midrule
\multirow{5}{*}{16}
& VdC            & $164.4\pm9.2$  & $126.18\pm6.53$  & $0.975\pm0.002$ & $0.316\pm0.040$ & $95.2\pm3.8$ \\
& Rand           & $161.6\pm13.0$ & $123.05\pm10.12$ & $0.976\pm0.004$ & $0.279\pm0.023$ & $95.4\pm3.1$ \\
& Anc ($p{=}1$)  & $154.8\pm9.5$  & $118.98\pm7.24$  & $0.976\pm0.002$ & $0.271\pm0.021$ & $94.8\pm2.3$ \\
& top-$p{=}0.95$ & $97.6\pm3.8$   & $75.33\pm3.11$   & $0.982\pm0.003$ & $0.391\pm0.032$ & $98.0\pm2.3$ \\
& top-$p{=}0.90$ & $65.9\pm7.6$   & $51.43\pm6.30$   & $0.983\pm0.002$ & $0.448\pm0.045$ & $98.9\pm2.0$ \\
\midrule
\multirow{5}{*}{32}
& VdC            & $356.6\pm24.1$ & $159.96\pm10.25$ & $0.966\pm0.002$ & $0.293\pm0.025$ & $99.3\pm0.8$ \\
& Rand           & $358.2\pm16.7$ & $159.43\pm8.10$  & $0.966\pm0.003$ & $0.281\pm0.014$ & $99.2\pm1.2$ \\
& Anc ($p{=}1$)  & $365.9\pm16.0$ & $162.06\pm6.40$  & $0.970\pm0.003$ & $0.275\pm0.020$ & $99.3\pm0.5$ \\
& top-$p{=}0.95$ & $251.9\pm10.7$ & $111.73\pm4.61$  & $0.976\pm0.003$ & $0.383\pm0.025$ & $99.9\pm0.2$ \\
& top-$p{=}0.90$ & $196.0\pm5.4$  & $87.50\pm2.57$   & $0.977\pm0.001$ & $0.440\pm0.021$ & $99.9\pm0.3$ \\
\bottomrule
\end{tabular}
\end{table}

\section{Complete Quantization Results Across Domains}
\label{sec:appendix-quant}

Tables~\ref{tab:qwen-all-domains-16}--\ref{tab:llama-all-domains-32} present complete concept graph statistics for all four evaluated domains (US Law, Git Version Control, Radiology Imaging, FAA Aviation) at both sequence lengths $\ell \in \{16, 32\}$. These extended results demonstrate that AWQ-4bit's superior concept coverage generalizes across structurally distinct knowledge domains.

\begin{table}[ht]
\caption{Concept graph statistics for Qwen2.5-7B-Instruct across all four domains at sequence length $\ell=16$.}
\label{tab:qwen-all-domains-16}
\begin{center}
\small
\begin{tabular}{llrrrrrr}
\toprule
Domain & Variant & Nodes & Edges & Density & Deg & Comp & CC\% \\
\midrule
\multirow{5}{*}{US Law}
& AWQ-4bit   & 1,052 & 2,736 & 0.0025 & 5.20 &  12 & 96.5 \\
& BF16       &   384 &   988 & 0.0067 & 5.15 &   8 & 95.3 \\
& GPTQ-Int8  &   371 &   915 & 0.0067 & 4.93 &   8 & 94.6 \\
& BNB-4bit   &   335 &   808 & 0.0072 & 4.82 &   8 & 93.7 \\
& GPTQ-Int4  &    55 &    85 & 0.0286 & 3.09 &   3 & 87.3 \\
\midrule
\multirow{5}{*}{Git}
& AWQ-4bit   &   157 &   652 & 0.0266 & 8.31 &   1 & 100.0 \\
& BNB-4bit   &   121 &   467 & 0.0322 & 7.72 &   1 & 100.0 \\
& GPTQ-Int8  &   117 &   443 & 0.0326 & 7.57 &   3 &  95.7 \\
& BF16       &   110 &   472 & 0.0394 & 8.58 &   1 & 100.0 \\
& GPTQ-Int4  &    93 &   316 & 0.0369 & 6.80 &   2 &  96.8 \\
\midrule
\multirow{5}{*}{Radiology}
& AWQ-4bit   &   665 & 2,370 & 0.0054 & 7.13 &   8 & 97.0 \\
& BF16       &   314 & 1,047 & 0.0107 & 6.67 &   4 & 97.5 \\
& BNB-4bit   &   289 &   937 & 0.0113 & 6.48 &  10 & 92.4 \\
& GPTQ-Int8  &   288 &   931 & 0.0113 & 6.47 &   4 & 96.5 \\
& GPTQ-Int4  &   118 &   301 & 0.0218 & 5.10 &   6 & 89.8 \\
\midrule
\multirow{5}{*}{FAA}
& AWQ-4bit   & 1,697 & 5,109 & 0.0018 & 6.02 &  14 & 98.1 \\
& BF16       & 1,085 & 3,343 & 0.0028 & 6.16 &  16 & 96.8 \\
& GPTQ-Int8  &   956 & 3,010 & 0.0033 & 6.30 &   6 & 98.7 \\
& BNB-4bit   &   693 & 1,800 & 0.0038 & 5.19 &   6 & 98.1 \\
& GPTQ-Int4  &   219 &   377 & 0.0079 & 3.44 &  12 & 85.8 \\
\bottomrule
\end{tabular}
\end{center}
\end{table}

\begin{table}[ht]
\caption{Concept graph statistics for Qwen2.5-7B-Instruct across all four domains at sequence length $\ell=32$.}
\label{tab:qwen-all-domains-32}
\begin{center}
\small
\begin{tabular}{llrrrrrr}
\toprule
Domain & Variant & Nodes & Edges & Density & Deg & Comp & CC\% \\
\midrule
\multirow{5}{*}{US Law}
& AWQ-4bit   & 1,899 & 7,408 & 0.0021 & 7.80 &   4 & 99.7 \\
& BF16       &   809 & 3,273 & 0.0050 & 8.09 &   6 & 97.4 \\
& GPTQ-Int8  &   786 & 3,132 & 0.0051 & 7.97 &   3 & 98.9 \\
& BNB-4bit   &   680 & 2,408 & 0.0052 & 7.08 &   5 & 98.5 \\
& GPTQ-Int4  &   101 &   211 & 0.0209 & 4.18 &   2 & 97.0 \\
\midrule
\multirow{5}{*}{Git}
& AWQ-4bit   &   526 & 2,962 & 0.0107 & 11.26 &   2 & 99.6 \\
& GPTQ-Int8  &   331 & 1,927 & 0.0176 & 11.64 &   2 & 98.2 \\
& BNB-4bit   &   311 & 1,783 & 0.0185 & 11.47 &   1 & 100.0 \\
& BF16       &   300 & 1,906 & 0.0212 & 12.71 &   1 & 100.0 \\
& GPTQ-Int4  &   206 & 1,144 & 0.0271 & 11.11 &   1 & 100.0 \\
\midrule
\multirow{5}{*}{Radiology}
& AWQ-4bit   & 1,674 & 8,793 & 0.0031 & 10.51 &   2 & 99.6 \\
& BF16       &   814 & 4,268 & 0.0065 & 10.49 &   1 & 100.0 \\
& GPTQ-Int8  &   785 & 3,940 & 0.0064 & 10.04 &   4 & 98.9 \\
& BNB-4bit   &   610 & 3,099 & 0.0083 & 10.16 &   3 & 98.4 \\
& GPTQ-Int4  &   267 &   883 & 0.0124 &  6.61 &   5 & 94.8 \\
\midrule
\multirow{5}{*}{FAA}
& AWQ-4bit   & 3,805 & 14,774 & 0.0010 & 7.77 &   9 & 99.3 \\
& BF16       & 2,002 &  7,759 & 0.0019 & 7.75 &   6 & 99.2 \\
& GPTQ-Int8  & 1,862 &  7,127 & 0.0021 & 7.66 &   7 & 98.7 \\
& BNB-4bit   & 1,217 &  3,802 & 0.0026 & 6.25 &  10 & 97.5 \\
& GPTQ-Int4  &   386 &    839 & 0.0057 & 4.35 &   4 & 95.6 \\
\bottomrule
\end{tabular}
\end{center}
\end{table}

\begin{table}[ht]
\caption{Concept graph statistics for Llama-3.1-8B-Instruct across all four domains at sequence length $\ell=16$.}
\label{tab:llama-all-domains-16}
\begin{center}
\small
\begin{tabular}{llrrrrrr}
\toprule
Domain & Variant & Nodes & Edges & Density & Deg & Comp & CC\% \\
\midrule
\multirow{5}{*}{US Law}
& AWQ-4bit   & 6,878 & 11,555 & 0.00024 & 3.36 & 601 & 80.9 \\
& BF16       & 5,259 & 10,841 & 0.00039 & 4.12 & 327 & 86.6 \\
& GPTQ-Int8  & 5,241 & 10,771 & 0.00039 & 4.11 & 349 & 85.8 \\
& BNB-4bit   & 4,029 &  7,524 & 0.00046 & 3.73 & 275 & 85.3 \\
& GPTQ-Int4  & 1,502 &  1,215 & 0.00054 & 1.62 & 391 & 40.8 \\
\midrule
\multirow{5}{*}{Git}
& AWQ-4bit   & 2,039 & 8,856 & 0.0021 & 8.69 &  14 & 98.4 \\
& GPTQ-Int8  & 1,698 & 8,044 & 0.0028 & 9.47 &   6 & 99.3 \\
& BF16       & 1,654 & 7,863 & 0.0029 & 9.51 &   7 & 99.0 \\
& BNB-4bit   & 1,585 & 6,980 & 0.0028 & 8.81 &   6 & 99.1 \\
& GPTQ-Int4  & 1,542 & 1,083 & 0.00046 & 1.40 & 485 & 29.1 \\
\midrule
\multirow{5}{*}{Radiology}
& AWQ-4bit   & 4,519 & 9,026 & 0.00044 & 3.99 & 326 & 84.5 \\
& BNB-4bit   & 3,851 & 8,267 & 0.00056 & 4.29 & 262 & 85.5 \\
& GPTQ-Int8  & 3,708 & 8,266 & 0.00060 & 4.46 & 189 & 89.3 \\
& BF16       & 3,659 & 8,180 & 0.00061 & 4.47 & 201 & 88.5 \\
& GPTQ-Int4  & 1,584 & 1,101 & 0.00044 & 1.39 & 502 & 27.0 \\
\midrule
\multirow{5}{*}{FAA}
& AWQ-4bit   & 6,795 & 12,076 & 0.00026 & 3.55 & 418 & 86.1 \\
& GPTQ-Int8  & 5,944 & 11,803 & 0.00033 & 3.97 & 312 & 88.5 \\
& BF16       & 5,791 & 11,694 & 0.00035 & 4.04 & 259 & 90.1 \\
& BNB-4bit   & 5,239 & 10,133 & 0.00037 & 3.87 & 292 & 87.7 \\
& GPTQ-Int4  & 1,547 &  1,091 & 0.00046 & 1.41 & 480 & 29.0 \\
\bottomrule
\end{tabular}
\end{center}
\end{table}

\begin{table}[ht]
\caption{Concept graph statistics for Llama-3.1-8B-Instruct across all four domains at sequence length $\ell=32$.}
\label{tab:llama-all-domains-32}
\begin{center}
\small
\begin{tabular}{llrrrrrr}
\toprule
Domain & Variant & Nodes & Edges & Density & Deg & Comp & CC\% \\
\midrule
\multirow{5}{*}{US Law}
& AWQ-4bit   & 14,654 & 34,439 & 0.00016 & 4.70 & 147 & 96.7 \\
& GPTQ-Int8  & 10,991 & 31,949 & 0.00026 & 5.81 &  73 & 97.9 \\
& BF16       & 10,922 & 32,220 & 0.00027 & 5.90 &  78 & 97.6 \\
& BNB-4bit   &  9,167 & 22,821 & 0.00027 & 4.98 & 103 & 96.4 \\
& GPTQ-Int4  &  5,360 &  4,730 & 0.00017 & 1.76 & 1,097 & 51.1 \\
\midrule
\multirow{5}{*}{Git}
& AWQ-4bit   & 3,914 & 21,734 & 0.0014 & 11.11 &   2 & 99.8 \\
& GPTQ-Int8  & 3,150 & 18,923 & 0.0019 & 12.01 &   4 & 99.5 \\
& BF16       & 3,139 & 18,696 & 0.0019 & 11.91 &   2 & 99.8 \\
& BNB-4bit   & 3,122 & 18,064 & 0.0019 & 11.57 &   4 & 99.4 \\
& GPTQ-Int4  & 4,131 &  3,151 & 0.00019 & 1.53 & 1,119 & 34.5 \\
\midrule
\multirow{5}{*}{Radiology}
& AWQ-4bit   & 9,510 & 27,707 & 0.00031 & 5.83 &  61 & 97.7 \\
& BNB-4bit   & 8,098 & 24,925 & 0.00038 & 6.16 &  69 & 97.4 \\
& GPTQ-Int8  & 7,323 & 24,616 & 0.00046 & 6.72 &  24 & 98.9 \\
& BF16       & 7,232 & 24,296 & 0.00047 & 6.72 &  28 & 98.8 \\
& GPTQ-Int4  & 4,620 &  3,596 & 0.00017 & 1.56 & 1,185 & 37.8 \\
\midrule
\multirow{5}{*}{FAA}
& AWQ-4bit   & 15,334 & 35,530 & 0.00015 & 4.63 & 101 & 97.8 \\
& GPTQ-Int8  & 12,881 & 34,042 & 0.00021 & 5.29 &  48 & 98.6 \\
& BF16       & 12,611 & 33,730 & 0.00021 & 5.35 &  45 & 98.7 \\
& BNB-4bit   & 11,701 & 30,132 & 0.00022 & 5.15 &  82 & 97.7 \\
& GPTQ-Int4  &  3,617 &  2,680 & 0.00021 & 1.48 & 1,019 & 30.6 \\
\bottomrule
\end{tabular}
\end{center}
\end{table}

\section{VdC Offset Robustness, Extended Results}
\label{sec:appendix-vdc}

\begin{table}[h]
\caption{Concept extraction stability across 8 VdC offsets at $\ell=16$. Pairwise Jaccard similarity and core concept percentage confirm robustness to quasi-random initialization. The quantization degradation pattern persists across all offsets.}
\label{tab:vdc-stability}
\centering\small
\begin{tabular}{lrrrr@{\quad}lrrrr}
\toprule
\multicolumn{5}{c}{\textbf{Qwen2.5-7B}} & \multicolumn{5}{c}{\textbf{Llama-3.1-8B}} \\
\cmidrule(r){1-5}\cmidrule(l){6-10}
Variant & Avg $N$ & Jacc. & Core & Core\% & Variant & Avg $N$ & Jacc. & Core & Core\% \\
\midrule
AWQ-4bit  & 428 & 29.0\% &  95 & 5.9\% & AWQ-4bit  & 2,334 & 15.4\% & 273 & 2.2\% \\
BF16      & 163 & 33.9\% &  42 & 7.6\% & BF16      & 1,838 & 18.5\% & 268 & 3.0\% \\
GPTQ-Int8 & 149 & 31.3\% &  43 & 8.3\% & GPTQ-Int8 & 1,842 & 18.1\% & 263 & 2.9\% \\
BNB-4bit  & 130 & 31.4\% &  30 & 6.5\% & BNB-4bit  & 1,398 & 16.8\% & 165 & 2.4\% \\
GPTQ-Int4 &  20 & 24.5\% &   4 & 4.6\% & GPTQ-Int4 &   430 &  7.3\% &  14 & 0.5\% \\
\bottomrule
\end{tabular}
\end{table}

Table~\ref{tab:vdc-extended} provides complete statistics for VdC offset experiments at $\ell=32$, complementing the $\ell=16$ results in the main text.

\begin{table}[ht]
\caption{VdC offset stability at $\ell=32$. Similar patterns to $\ell=16$ confirm robustness across sequence lengths.}
\label{tab:vdc-extended}
\begin{center}
\small
\begin{tabular}{lrrrr@{\hskip 0.3in}lrrrr}
\toprule
\multicolumn{5}{c}{\textbf{Qwen2.5-7B-Instruct}} & \multicolumn{5}{c}{\textbf{Llama-3.1-8B-Instruct}} \\
\cmidrule(r){1-5} \cmidrule(l){6-10}
Variant & Avg $N$ & Jaccard & Core & Core\% & Variant & Avg $N$ & Jaccard & Core & Core\% \\
\midrule
AWQ-4bit   & 793 & 26.5\% & 152 & 4.9\% & AWQ-4bit   & 4,867 & 13.3\% & 439 & 1.6\% \\
BF16       & 355 & 30.6\% &  84 & 6.6\% & BF16       & 3,822 & 15.8\% & 416 & 2.1\% \\
GPTQ-Int8  & 332 & 29.9\% &  78 & 6.5\% & GPTQ-Int8  & 3,809 & 16.1\% & 430 & 2.2\% \\
BNB-4bit   & 266 & 29.6\% &  56 & 5.7\% & BNB-4bit   & 3,092 & 14.7\% & 295 & 1.8\% \\
GPTQ-Int4  &  39 & 26.5\% &   9 & 5.5\% & GPTQ-Int4  & 1,531 &  5.9\% &  47 & 0.5\% \\
\bottomrule
\end{tabular}
\end{center}
\end{table}

\section{Pairwise Concept Overlap Between Quantization Methods}
\label{sec:appendix-overlap}

To assess whether different quantization methods extract similar or divergent concept sets, we compute pairwise Jaccard similarity coefficients between concept sets produced by all quantization variant pairs, within each fixed model-domain-generation-length configuration. Tables~\ref{tab:overlap-qwen} and~\ref{tab:overlap-llama} report representative results for the US Law domain at generation length $\ell = 16$.

\begin{table}[h]
\caption{Pairwise concept overlap for Qwen2.5-7B-Instruct on the US Law domain at generation length $\ell = 16$.}
\label{tab:overlap-qwen}
\centering
\small
\begin{tabular}{lrrr}
\toprule
Variant Pair & Overlap & Jaccard & Nodes (A, B) \\
\midrule
BF16 / GPTQ-Int8      & 168 & 28.6\% & (384, 371) \\
BF16 / BNB-4bit       & 148 & 25.9\% & (384, 335) \\
GPTQ-Int8 / BNB-4bit  & 135 & 23.6\% & (371, 335) \\
\midrule
AWQ-4bit / BF16       & 215 & 17.6\% & (1,052, 384) \\
AWQ-4bit / GPTQ-Int8  & 182 & 14.7\% & (1,052, 371) \\
AWQ-4bit / BNB-4bit   & 181 & 15.0\% & (1,052, 335) \\
\midrule
AWQ-4bit / GPTQ-Int4  &  22 & 2.0\%  & (1,052, 55) \\
BF16 / GPTQ-Int4      &  25 & 6.0\%  & (384, 55) \\
\midrule
\textit{All 5 variants} & \textit{15} & \textit{N/A} & \textit{N/A} \\
\bottomrule
\end{tabular}
\end{table}

\begin{table}[h]
\caption{Pairwise concept overlap for Llama-3.1-8B-Instruct on the US Law domain at generation length $\ell = 16$. GPTQ-Int4 exhibits near-complete divergence from all other quantization methods.}
\label{tab:overlap-llama}
\centering
\small
\begin{tabular}{lrrr}
\toprule
Variant Pair & Overlap & Jaccard & Nodes (A, B) \\
\midrule
BF16 / GPTQ-Int8       & 1,522 & 17.0\% & (5,259, 5,241) \\
AWQ-4bit / BF16        & 1,407 & 13.1\% & (6,878, 5,259) \\
AWQ-4bit / GPTQ-Int8   & 1,366 & 12.7\% & (6,878, 5,241) \\
BF16 / BNB-4bit        & 1,061 & 12.9\% & (5,259, 4,029) \\
AWQ-4bit / BNB-4bit    & 1,023 & 10.4\% & (6,878, 4,029) \\
\midrule
AWQ-4bit / GPTQ-Int4   & 5 & 0.1\% & (6,878, 1,502) \\
BF16 / GPTQ-Int4       & 4 & 0.1\% & (5,259, 1,502) \\
GPTQ-Int8 / GPTQ-Int4  & 3 & 0.0\% & (5,241, 1,502) \\
BNB-4bit / GPTQ-Int4   & 1 & 0.0\% & (4,029, 1,502) \\
\midrule
\textit{All 5 variants} & \textit{0} & \textit{N/A} & \textit{N/A} \\
\bottomrule
\end{tabular}
\end{table}

\section{Model Sources and Extraction Prompts}
\label{sec:appendix-model-prompts}

\subsection{Quantized Model Sources}

All quantized models were obtained from the Hugging Face Model Hub. Table~\ref{tab:model-sources} lists the exact repository paths used for each model variant in our quantization experiments.

\begin{table}[h]
\caption{Model repository sources for quantization experiments.}
\label{tab:model-sources}
\centering
\small
\begin{tabular}{ll}
\toprule
Variant & Repository \\
\midrule
\multicolumn{2}{c}{\textit{Qwen2.5-7B-Instruct}} \\
\midrule
BF16      & \texttt{Qwen/Qwen2.5-7B-Instruct} \\
GPTQ-Int8 & \texttt{Qwen/Qwen2.5-7B-Instruct-GPTQ-Int8} \\
AWQ-4bit  & \texttt{Qwen/Qwen2.5-7B-Instruct-AWQ} \\
GPTQ-Int4 & \texttt{Qwen/Qwen2.5-7B-Instruct-GPTQ-Int4} \\
BNB-4bit  & \texttt{unsloth/Qwen2.5-7B-Instruct-bnb-4bit} \\
\midrule
\multicolumn{2}{c}{\textit{Llama-3.1-8B-Instruct}} \\
\midrule
BF16      & \texttt{meta-llama/Llama-3.1-8B-Instruct} \\
GPTQ-Int8 & \texttt{abdo-Mansour/Meta-Llama-3.1-8B-Instruct-GPTQ-8bit} \\
AWQ-4bit  & \texttt{hugging-quants/Meta-Llama-3.1-8B-Instruct-AWQ-INT4} \\
GPTQ-Int4 & \texttt{hugging-quants/Meta-Llama-3.1-8B-Instruct-GPTQ-INT4} \\
BNB-4bit  & \texttt{unsloth/Meta-Llama-3.1-8B-Instruct-bnb-4bit} \\
\bottomrule
\end{tabular}
\end{table}

All models were loaded using the Transformers library. For BF16 baselines, we used \texttt{torch.bfloat16}, while quantized variants were loaded with \texttt{dtype="auto"} to allow backend-specific precision handling. BNB-4bit models rely on the \texttt{bitsandbytes} library as packaged by Unsloth.

\subsection{Domain-Specific Extraction Prompts}
\label{sec:extraction-prompts}

All prompts were delivered as single-turn user messages via the model's chat template (\texttt{apply\_chat\_template} with \texttt{add\_generation\_prompt=True}). No few-shot examples were provided. The exact prompts used in each experiment are listed below verbatim.

\paragraph{Experiments 1--3, 5, 7, 8 (quantization and saturation, US Law).}
\begin{quote}
\small\ttfamily
Generate United States bar exam legal concepts as keywords.\\
\\
Please output ONE concept per line.\\
Each concept can be multiple words if needed.\\
Do not include explanations or extra text.\\
Please begin from any random concept.\\
Please use English.
\end{quote}

\paragraph{Experiments 1--2 (other domains).}
\begin{description}
\item[Git:] \textit{Generate Git and version control concepts as keywords. [+ same format instructions]}
\item[Radiology:] \textit{Generate radiology and medical imaging concepts as keywords. [+ same format instructions]}
\item[FAA:] \textit{Generate FAA aviation flight rules and airspace concepts as keywords. [+ same format instructions]}
\end{description}

\paragraph{Experiment 4 (sampling baselines) and Experiment 9 (top-$p$).}
Same as US Law prompt above.

\paragraph{Experiment 6 (prompt paraphrase robustness).} Five variants, all requesting legal \emph{concepts}, differing only in surface phrasing:
\begin{description}
\item[Baseline:] Same as US Law prompt above.
\item[List relevant:] \textit{List legal concepts relevant to the United States bar examination. Output one concept per line. Each concept may contain multiple words. Do not include explanations or commentary. Please begin from any random concept. Use English.}
\item[Study materials:] \textit{Produce concepts commonly appearing in United States bar exam materials. Write one concept per line. Multi-word concepts are allowed. Avoid explanations and additional text. Please begin from any random concept. Use English.}
\item[Concise:] \textit{Output United States bar exam legal concepts. One concept per line. Concepts may be single words or multi-word phrases. No explanations. Begin from any random concept. English only.}
\item[Study concepts:] \textit{Generate legal concepts studied for the United States bar exam. Return exactly one concept per line. A concept may consist of multiple words. Do not add explanations, numbering, or extra text. Start from any random concept. Use English.}
\end{description}

\paragraph{Experiment 3 (hallucination, 21 models).} Same as US Law prompt above, applied identically across all 21 models.

\subsection{Arithmetic Decoding Implementation Details}
\label{sec:arithmetic-impl}

\paragraph{Bijection and code assignment.} We use arithmetic sampling \citep{vilnis2023arithmeticsamplingparalleldiverse}, which implements a bijection between sequences and codes $z \in [0,1)$. Given a code $z$, decoding proceeds greedily: at each step $t$, the interval $[L_t, U_t)$ is subdivided by the current token's CDF, and the token whose sub-interval contains $z$ is selected. The code uniquely determines the sequence without any additional randomness.

\paragraph{EOS and length handling.} Generation terminates when (a) the selected token is the EOS token, or (b) the maximum length $\ell \in \{16, 32\}$ is reached. If EOS is selected before $\ell$ tokens, the sequence ends there; the final incomplete line (no trailing newline) is discarded during concept extraction to avoid partial-concept artifacts.

\paragraph{Tokenizer and precision.} All experiments use fast tokenizers (\texttt{use\_fast=True}). CDF computations use float32 logits from the model's forward pass; no top-$k$ or top-$p$ filtering is applied during arithmetic decoding (the full vocabulary CDF is used). For GPTQ-Int4 Llama, \texttt{desc\_act=False} is set to avoid activation-order errors with the AutoGPTQ backend.

\paragraph{Determinism.} Within a single GPU and software stack, results are bitwise-deterministic: the VdC code fully determines the decoding path, and we set \texttt{torch.backends.cudnn.deterministic=True} and \texttt{benchmark=False}. Floating-point non-associativity across different GPU architectures or CUDA versions may produce minor token-probability differences; we do not claim bitwise cross-hardware reproducibility, but the statistical properties (coverage, Jaccard) are robust to such variation as evidenced by the VdC offset robustness experiment (Section~\ref{sec:exp2-vdc-offsets}).

\paragraph{Perplexity measurement.} Perplexity is computed as token-level cross-entropy on the extraction prompt itself (the user-turn text only, not the assistant response), averaged over tokens, then exponentiated. This is a standard input-conditioned perplexity measurement, not an explanation-level or continuation-level measure. It reflects how the model assigns probability to the prompt tokens under the chat template formatting, providing a reference signal that is independent of generation behavior.

\paragraph{AWQ entropy note.} A natural question is whether AWQ's higher concept coverage reflects genuine preservation of long-tail knowledge versus a broadening of the output distribution (higher entropy). Our normalized coverage metrics (Cov/10kT, Cov/1kL) partially address this by controlling for output length, and the results show that a normalized advantage persists for Llama but is substantially reduced for Qwen. Fully isolating knowledge preservation from entropy change would require temperature-matching across variants , a direction we leave to future work, noting that our framing of concept coverage as a property of the model/prompt/decoder triad explicitly includes distributional entropy as a component of that triad.

\section{Sampled Legal Concepts}
\label{sec:sampled-concepts}

This appendix lists all 200 concepts sampled from the top 2 MMLU-Pro Law models (google/gemma-2-9b-it and mistralai/Mistral-Nemo-Instruct-2407), 
stratified by extraction frequency. Concepts marked with \checkmark{} were verified 
in CourtListener's legal database and those marked with a cross were not.

\subsection{gemma-2-9b-it}
\label{sec:concepts-google-gemma-2-9b-it}

Sampled 199 concepts from google/gemma-2-9b-it. 
Verified: 175 (87.9\%), 
Hallucinated: 24 (12.1\%).

\paragraph{High Frequency Concepts (Top 25\%, n=49):}

\begin{multicols}{2}
\begin{itemize}[leftmargin=*, itemsep=0pt, parsep=2pt]
  \item [\checkmark] \texttt{statue of limitation} (\small freq=440)
  \item [\checkmark] \texttt{negligence per se} (\small freq=369)
  \item [\checkmark] \texttt{miranda right} (\small freq=362)
  \item [\checkmark] \texttt{res judicata} (\small freq=361)
  \item [\checkmark] \texttt{statutory interpretation} (\small freq=292)
  \item [\checkmark] \texttt{standing} (\small freq=255)
  \item [\checkmark] \texttt{due process clause} (\small freq=246)
  \item [\checkmark] \texttt{contractformation} (\small freq=213)
  \item [\checkmark] \texttt{negligence} (\small freq=198)
  \item [\checkmark] \texttt{tort} (\small freq=196)
  \item [\checkmark] \texttt{res ipsa loquitur} (\small freq=179)
  \item [\checkmark] \texttt{due proces} (\small freq=175)
  \item [\checkmark] \texttt{burden of proof} (\small freq=158)
  \item [\checkmark] \texttt{mens rea} (\small freq=154)
  \item [\checkmark] \texttt{battery} (\small freq=151)
  \item [\checkmark] \texttt{contitutional law} (\small freq=146)
  \item [\checkmark] \texttt{strict liability} (\small freq=146)
  \item [\checkmark] \texttt{jurisdiction} (\small freq=127)
  \item [\checkmark] \texttt{respondeat superior} (\small freq=123)
  \item [\checkmark] \texttt{miranda warning} (\small freq=117)
  \item [\checkmark] \texttt{precedent} (\small freq=116)
  \item [\checkmark] \texttt{civl procedure} (\small freq=113)
  \item [\checkmark] \texttt{federal rule of evidence} (\small freq=112)
  \item [\checkmark] \texttt{statue of fraud} (\small freq=103)
  \item [\checkmark] \texttt{informed consent} (\small freq=97)
  \item [\checkmark] \texttt{offer and acceptance} (\small freq=95)
  \item [\checkmark] \texttt{product liability} (\small freq=88)
  \item [\checkmark] \texttt{standard of proof} (\small freq=83)
  \item [\checkmark] \texttt{first amendment right} (\small freq=81)
  \item [\checkmark] \texttt{legal malpractice} (\small freq=76)
  \item [\checkmark] \texttt{contract} (\small freq=73)
  \item [\checkmark] \texttt{hearsay evidence} (\small freq=73)
  \item [\checkmark] \texttt{discovery} (\small freq=71)
  \item [\checkmark] \texttt{consideration} (\small freq=70)
  \item [\checkmark] \texttt{contract law} (\small freq=70)
  \item [\checkmark] \texttt{estoppel} (\small freq=70)
  \item[$\times$] \texttt{tortsnegligence} (\small freq=69)
  \item [\checkmark] \texttt{hearsay exception} (\small freq=68)
  \item [\checkmark] \texttt{habeas corpu} (\small freq=67)
  \item [\checkmark] \texttt{legal precedent} (\small freq=65)
  \item [\checkmark] \texttt{proximate cause} (\small freq=65)
  \item [\checkmark] \texttt{duty of care} (\small freq=61)
  \item [\checkmark] \texttt{secured transaction} (\small freq=59)
  \item [\checkmark] \texttt{damage} (\small freq=57)
  \item [\checkmark] \texttt{standing to sue} (\small freq=57)
  \item [\checkmark] \texttt{federal jurisdiction} (\small freq=55)
  \item [\checkmark] \texttt{stare decisi} (\small freq=55)
  \item [\checkmark] \texttt{rule against perpetuity} (\small freq=54)
  \item [\checkmark] \texttt{causation} (\small freq=54)
\end{itemize}
\end{multicols}

\paragraph{Medium Frequency Concepts (Middle 50\%, n=100):}

\begin{multicols}{2}
\begin{itemize}[leftmargin=*, itemsep=0pt, parsep=2pt]
  \item [\checkmark] \texttt{fourth amendment} (\small freq=54)
  \item [\checkmark] \texttt{jurisprudence} (\small freq=6)
  \item [\checkmark] \texttt{expert testimony} (\small freq=5)
  \item [\checkmark] \texttt{ignorance of law} (\small freq=5)
  \item [\checkmark] \texttt{fraudulent inducement} (\small freq=5)
  \item [\checkmark] \texttt{murder} (\small freq=5)
  \item [\checkmark] \texttt{establishment clause} (\small freq=5)
  \item [\checkmark] \texttt{magna carta} (\small freq=5)
  \item [\checkmark] \texttt{cruel and unusual punishment} (\small freq=4)
  \item [\checkmark] \texttt{civil litigation} (\small freq=4)
  \item [\checkmark] \texttt{amicus curiae brief} (\small freq=4)
  \item [\checkmark] \texttt{appellate jurisdiction} (\small freq=4)
  \item [\checkmark] \texttt{equitable tolling} (\small freq=4)
  \item [\checkmark] \texttt{last clear chance doctrine} (\small freq=4)
  \item [\checkmark] \texttt{evidentiary hearsay exception} (\small freq=4)
  \item [\checkmark] \texttt{duty of confidentiality} (\small freq=4)
  \item [\checkmark] \texttt{discovery motion} (\small freq=4)
  \item [\checkmark] \texttt{federal procedure} (\small freq=4)
  \item [\checkmark] \texttt{civil procedure discovery} (\small freq=4)
  \item [\checkmark] \texttt{united states constitution} (\small freq=3)
  \item [\checkmark] \texttt{daubert standard} (\small freq=3)
  \item [\checkmark] \texttt{negligence claim} (\small freq=3)
  \item [\checkmark] \texttt{intent to defraud} (\small freq=3)
  \item [\checkmark] \texttt{miranda v. arizona} (\small freq=3)
  \item [\checkmark] \texttt{foreseeability of harm} (\small freq=3)
  \item [\checkmark] \texttt{right to remain silent} (\small freq=3)
  \item [\checkmark] \texttt{memorandum of law} (\small freq=3)
  \item [\checkmark] \texttt{negligence duty of care} (\small freq=3)
  \item[$\times$] \texttt{torts intentional tort} (\small freq=3)
  \item [\checkmark] \texttt{third party beneficiary} (\small freq=3)
  \item[$\times$] \texttt{pleading special defense} (\small freq=3)
  \item [\checkmark] \texttt{takings clause} (\small freq=3)
  \item [\checkmark] \texttt{in rem jurisdiction} (\small freq=3)
  \item [\checkmark] \texttt{estate planning} (\small freq=3)
  \item [\checkmark] \texttt{sufficiency of the indictment} (\small freq=2)
  \item [\checkmark] \texttt{alternative pleading} (\small freq=2)
  \item [\checkmark] \texttt{exceptions to hearsay} (\small freq=2)
  \item [\checkmark] \texttt{real property law} (\small freq=2)
  \item [\checkmark] \texttt{battery negligence} (\small freq=2)
  \item [\checkmark] \texttt{consideration in contract formation} (\small freq=2)
  \item [\checkmark] \texttt{federalist paper} (\small freq=2)
  \item[$\times$] \texttt{protoimpl} (\small freq=2)
  \item [\checkmark] \texttt{pleading special damage} (\small freq=2)
  \item [\checkmark] \texttt{evidentiary admissibility} (\small freq=2)
  \item [\checkmark] \texttt{legal intent} (\small freq=2)
  \item [\checkmark] \texttt{evidence hearsay} (\small freq=2)
  \item[$\times$] \texttt{negligent infliction emotional distres} (\small freq=2)
  \item [\checkmark] \texttt{ultra vires doctrine} (\small freq=2)
  \item [\checkmark] \texttt{california civil code} (\small freq=2)
  \item [\checkmark] \texttt{negotiation and settlement} (\small freq=2)
  \item [\checkmark] \texttt{slander per se} (\small freq=2)
  \item [\checkmark] \texttt{treaties and convention} (\small freq=2)
  \item [\checkmark] \texttt{statutory authority} (\small freq=2)
  \item [\checkmark] \texttt{fair use} (\small freq=2)
  \item [\checkmark] \texttt{slander and libel} (\small freq=2)
  \item [\checkmark] \texttt{common law remedy} (\small freq=2)
  \item [\checkmark] \texttt{common law tort} (\small freq=2)
  \item [\checkmark] \texttt{agency and principal} (\small freq=2)
  \item [\checkmark] \texttt{mock trial} (\small freq=2)
  \item [\checkmark] \texttt{due process fourteenth amendment} (\small freq=2)
  \item[$\times$] \texttt{standing and mootnes} (\small freq=2)
  \item [\checkmark] \texttt{free speech limitation} (\small freq=2)
  \item [\checkmark] \texttt{business association} (\small freq=2)
  \item[$\times$] \texttt{yoksa} (\small freq=2)
  \item[$\times$] \texttt{legal analysi} (\small freq=2)
  \item [\checkmark] \texttt{declaratory judgment} (\small freq=2)
  \item [\checkmark] \texttt{intro evidence} (\small freq=2)
  \item [\checkmark] \texttt{trust law} (\small freq=2)
  \item [\checkmark] \texttt{negotiation and bargaining} (\small freq=2)
  \item [\checkmark] \texttt{higher education} (\small freq=2)
  \item [\checkmark] \texttt{innocent until proven guilty} (\small freq=2)
  \item [\checkmark] \texttt{relevance and materiality} (\small freq=2)
  \item [\checkmark] \texttt{plea bargain} (\small freq=2)
  \item [\checkmark] \texttt{maritime law} (\small freq=2)
  \item [\checkmark] \texttt{sovereign immunity} (\small freq=2)
  \item [\checkmark] \texttt{injunctive relief} (\small freq=2)
  \item [\checkmark] \texttt{valid contract formation} (\small freq=2)
  \item [\checkmark] \texttt{learned hand rule} (\small freq=2)
  \item [\checkmark] \texttt{tort law negligence} (\small freq=2)
  \item[$\times$] \texttt{vazquez accident} (\small freq=2)
  \item [\checkmark] \texttt{reasonableness standard} (\small freq=2)
  \item [\checkmark] \texttt{irreconcilable difference} (\small freq=2)
  \item [\checkmark] \texttt{monopolistic competition} (\small freq=2)
  \item [\checkmark] \texttt{demand for jury trial} (\small freq=2)
  \item [\checkmark] \texttt{lochner era} (\small freq=2)
  \item [\checkmark] \texttt{interlocutory appeal} (\small freq=1)
  \item [\checkmark] \texttt{oaths and affirmation} (\small freq=1)
  \item[$\times$] \texttt{legal profession ethic} (\small freq=1)
  \item [\checkmark] \texttt{subpoena} (\small freq=1)
  \item [\checkmark] \texttt{right to bear arm} (\small freq=1)
  \item[$\times$] \texttt{notice and actual malice} (\small freq=1)
  \item [\checkmark] \texttt{automobile negligence} (\small freq=1)
  \item [\checkmark] \texttt{tort reform} (\small freq=1)
  \item [\checkmark] \texttt{entrapment defense} (\small freq=1)
  \item [\checkmark] \texttt{hague convention on child abduction} (\small freq=1)
  \item [\checkmark] \texttt{mortgage} (\small freq=1)
  \item[$\times$] \texttt{habitual drunkennes} (\small freq=1)
  \item [\checkmark] \texttt{cause of action} (\small freq=1)
  \item [\checkmark] \texttt{unjust enrichment} (\small freq=1)
  \item [\checkmark] \texttt{immunization requirement} (\small freq=1)
\end{itemize}
\end{multicols}

\paragraph{Low Frequency Concepts (Bottom 25\%, n=50):}

\begin{multicols}{2}
\begin{itemize}[leftmargin=*, itemsep=0pt, parsep=2pt]
  \item [\checkmark] \texttt{standing arbitration agreement} (\small freq=1)
  \item [\checkmark] \texttt{jurisdiction long arm statute} (\small freq=1)
  \item [\checkmark] \texttt{probation} (\small freq=1)
  \item [\checkmark] \texttt{anticipatory breach of contract} (\small freq=1)
  \item[$\times$] \texttt{introversion and extroversion} (\small freq=1)
  \item [\checkmark] \texttt{constitutionality of statute} (\small freq=1)
  \item[$\times$] \texttt{assuming the validity of a contract} (\small freq=1)
  \item [\checkmark] \texttt{accomplice testimony} (\small freq=1)
  \item [\checkmark] \texttt{case law precedent} (\small freq=1)
  \item [\checkmark] \texttt{jury} (\small freq=1)
  \item[$\times$] \texttt{stretting resource} (\small freq=1)
  \item [\checkmark] \texttt{constitution} (\small freq=1)
  \item [\checkmark] \texttt{civil conspiracy} (\small freq=1)
  \item [\checkmark] \texttt{freedom of contract} (\small freq=1)
  \item[$\times$] \texttt{fourth amendment searches and seizure} (\small freq=1)
  \item[$\times$] \texttt{torts duty of care} (\small freq=1)
  \item [\checkmark] \texttt{negligence <br>} (\small freq=1)
  \item[$\times$] \texttt{kantian ethic} (\small freq=1)
  \item [\checkmark] \texttt{no contest clause} (\small freq=1)
  \item [\checkmark] \texttt{prosecutorial misconduct} (\small freq=1)
  \item [\checkmark] \texttt{admission} (\small freq=1)
  \item [\checkmark] \texttt{exclusive jurisdiction} (\small freq=1)
  \item[$\times$] \texttt{moneyservice business act} (\small freq=1)
  \item[$\times$] \texttt{malleable interest} (\small freq=1)
  \item [\checkmark] \texttt{appellate procedure} (\small freq=1)
  \item [\checkmark] \texttt{public policy exception} (\small freq=1)
  \item [\checkmark] \texttt{elements of contract formation} (\small freq=1)
  \item [\checkmark] \texttt{substantial performance} (\small freq=1)
  \item [\checkmark] \texttt{res judicata doctrine} (\small freq=1)
  \item [\checkmark] \texttt{rico statute} (\small freq=1)
  \item [\checkmark] \texttt{severability clause} (\small freq=1)
  \item [\checkmark] \texttt{jury instruction} (\small freq=1)
  \item [\checkmark] \texttt{state bar exam} (\small freq=1)
  \item[$\times$] \texttt{conversion causation in fact standing} (\small freq=1)
  \item [\checkmark] \texttt{duress and coercion} (\small freq=1)
  \item [\checkmark] \texttt{equity and good conscience} (\small freq=1)
  \item [\checkmark] \texttt{legal fiduciary} (\small freq=1)
  \item [\checkmark] \texttt{constitutional privilege} (\small freq=1)
  \item [\checkmark] \texttt{constructive notice} (\small freq=1)
  \item[$\times$] \texttt{cost benefit analysi} (\small freq=1)
  \item [\checkmark] \texttt{contingency contract} (\small freq=1)
  \item [\checkmark] \texttt{retained earning} (\small freq=1)
  \item [\checkmark] \texttt{heir} (\small freq=1)
  \item [\checkmark] \texttt{affirmed and reversed} (\small freq=1)
  \item [\checkmark] \texttt{agency theory} (\small freq=1)
  \item [\checkmark] \texttt{stand your ground} (\small freq=1)
  \item[$\times$] \texttt{olahraga professional liability} (\small freq=1)
  \item [\checkmark] \texttt{consent to treatment} (\small freq=1)
  \item[$\times$] \texttt{privilege based immunity} (\small freq=1)
  \item [\checkmark] \texttt{duress and undue influence} (\small freq=1)
\end{itemize}
\end{multicols}

\subsection{Mistral-Nemo-Instruct-2407}
\label{sec:concepts-mistralai-Mistral-Nemo-Instruct-2407}

Sampled 197 concepts from mistralai/Mistral-Nemo-Instruct-2407. 
Verified: 159 (80.7\%), 
Hallucinated: 38 (19.3\%).

\paragraph{High Frequency Concepts (Top 25\%, n=49):}

\begin{multicols}{2}
\begin{itemize}[leftmargin=*, itemsep=0pt, parsep=2pt]
  \item [\checkmark] \texttt{contractformation} (\small freq=204)
  \item [\checkmark] \texttt{burden of proof} (\small freq=197)
  \item [\checkmark] \texttt{constitution law} (\small freq=184)
  \item [\checkmark] \texttt{hearsay} (\small freq=178)
  \item [\checkmark] \texttt{civilprocedure} (\small freq=139)
  \item [\checkmark] \texttt{negligence} (\small freq=121)
  \item [\checkmark] \texttt{adverse possession} (\small freq=110)
  \item [\checkmark] \texttt{consideration} (\small freq=104)
  \item [\checkmark] \texttt{miranda warning} (\small freq=101)
  \item [\checkmark] \texttt{tortiousinterference} (\small freq=92)
  \item [\checkmark] \texttt{tort} (\small freq=91)
  \item [\checkmark] \texttt{due proces} (\small freq=83)
  \item [\checkmark] \texttt{criminalprocedure} (\small freq=78)
  \item [\checkmark] \texttt{administrative law} (\small freq=76)
  \item [\checkmark] \texttt{misdemeanor} (\small freq=74)
  \item [\checkmark] \texttt{1. hearsay} (\small freq=71)
  \item [\checkmark] \texttt{criminallaw} (\small freq=69)
  \item [\checkmark] \texttt{unjust enrichment} (\small freq=62)
  \item [\checkmark] \texttt{contractlaw} (\small freq=61)
  \item [\checkmark] \texttt{1. civil procedure} (\small freq=51)
  \item [\checkmark] \texttt{* constitutional law} (\small freq=51)
  \item [\checkmark] \texttt{rule against perpetuity} (\small freq=51)
  \item [\checkmark] \texttt{attorney client privilege} (\small freq=50)
  \item [\checkmark] \texttt{estoppel} (\small freq=43)
  \item [\checkmark] \texttt{batter} (\small freq=37)
  \item [\checkmark] \texttt{offer and acceptance} (\small freq=36)
  \item [\checkmark] \texttt{miranda right} (\small freq=36)
  \item [\checkmark] \texttt{contract interpreting} (\small freq=36)
  \item [\checkmark] \texttt{privity of contract} (\small freq=35)
  \item [\checkmark] \texttt{vicarious liability} (\small freq=35)
  \item [\checkmark] \texttt{res judicata} (\small freq=34)
  \item [\checkmark] \texttt{agency} (\small freq=33)
  \item [\checkmark] \texttt{negligence per se} (\small freq=31)
  \item [\checkmark] \texttt{tortlaw} (\small freq=30)
  \item [\checkmark] \texttt{contract} (\small freq=30)
  \item [\checkmark] \texttt{1. contract formation} (\small freq=30)
  \item [\checkmark] \texttt{first amendment} (\small freq=29)
  \item [\checkmark] \texttt{causation} (\small freq=27)
  \item [\checkmark] \texttt{real property} (\small freq=27)
  \item [\checkmark] \texttt{unconscionability} (\small freq=26)
  \item [\checkmark] \texttt{adversary system} (\small freq=26)
  \item [\checkmark] \texttt{contract remedy} (\small freq=25)
  \item [\checkmark] \texttt{contract ucc} (\small freq=25)
  \item [\checkmark] \texttt{prima facie case} (\small freq=25)
  \item[$\times$] \texttt{2. due proces} (\small freq=24)
  \item [\checkmark] \texttt{1. negligence} (\small freq=23)
  \item [\checkmark] \texttt{jurisdiction} (\small freq=23)
  \item [\checkmark] \texttt{mens rea} (\small freq=23)
  \item [\checkmark] \texttt{duty of care} (\small freq=22)
\end{itemize}
\end{multicols}

\paragraph{Medium Frequency Concepts (Middle 50\%, n=98):}

\begin{multicols}{2}
\begin{itemize}[leftmargin=*, itemsep=0pt, parsep=2pt]
  \item [\checkmark] \texttt{evidence} (\small freq=22)
  \item [\checkmark] \texttt{contested divorce} (\small freq=2)
  \item [\checkmark] \texttt{state criminal law} (\small freq=2)
  \item [\checkmark] \texttt{ancillary doctrine} (\small freq=2)
  \item [\checkmark] \texttt{civ pro} (\small freq=2)
  \item [\checkmark] \texttt{assume} (\small freq=2)
  \item [\checkmark] \texttt{assignment} (\small freq=2)
  \item[$\times$] \texttt{2. substantive due proces} (\small freq=2)
  \item [\checkmark] \texttt{plaintiff} (\small freq=2)
  \item [\checkmark] \texttt{unreasonable search and seizure} (\small freq=2)
  \item [\checkmark] \texttt{estate planning} (\small freq=2)
  \item [\checkmark] \texttt{3. tort} (\small freq=2)
  \item [\checkmark] \texttt{prevailing party} (\small freq=2)
  \item [\checkmark] \texttt{mistake of fact} (\small freq=2)
  \item[$\times$] \texttt{bar exam keyword} (\small freq=2)
  \item [\checkmark] \texttt{**commerce clause**} (\small freq=2)
  \item [\checkmark] \texttt{civil conspiracy} (\small freq=2)
  \item [\checkmark] \texttt{territorial jurisdiction} (\small freq=2)
  \item [\checkmark] \texttt{affirmative action} (\small freq=2)
  \item [\checkmark] \texttt{clear and present danger} (\small freq=2)
  \item [\checkmark] \texttt{state action doctrine} (\small freq=2)
  \item [\checkmark] \texttt{quasi contract} (\small freq=2)
  \item [\checkmark] \texttt{ship bottom} (\small freq=2)
  \item [\checkmark] \texttt{expert testimony} (\small freq=2)
  \item [\checkmark] \texttt{assume the fact} (\small freq=2)
  \item[$\times$] \texttt{municipal validity} (\small freq=2)
  \item [\checkmark] \texttt{emotional distres} (\small freq=2)
  \item [\checkmark] \texttt{proton} (\small freq=2)
  \item [\checkmark] \texttt{release} (\small freq=2)
  \item [\checkmark] \texttt{* tort law} (\small freq=2)
  \item [\checkmark] \texttt{volenti non fit injuria} (\small freq=2)
  \item [\checkmark] \texttt{criminal responsibility} (\small freq=2)
  \item [\checkmark] \texttt{bar membership} (\small freq=2)
  \item [\checkmark] \texttt{arguable error} (\small freq=2)
  \item [\checkmark] \texttt{real estate} (\small freq=2)
  \item [\checkmark] \texttt{conspiracy law} (\small freq=2)
  \item [\checkmark] \texttt{formal offer and acceptance} (\small freq=2)
  \item [\checkmark] \texttt{title theory} (\small freq=2)
  \item [\checkmark] \texttt{legal realism} (\small freq=2)
  \item[$\times$] \texttt{mixed wildlife} (\small freq=2)
  \item [\checkmark] \texttt{proof burden} (\small freq=2)
  \item [\checkmark] \texttt{waiver} (\small freq=2)
  \item [\checkmark] \texttt{renvoi} (\small freq=2)
  \item [\checkmark] \texttt{**procedural due process**} (\small freq=1)
  \item [\checkmark] \texttt{understanding of american law} (\small freq=1)
  \item[$\times$] \texttt{1. **civil burden of proof**} (\small freq=1)
  \item[$\times$] \texttt{1. bifurcation of proceedings} (\small freq=1)
  \item [\checkmark] \texttt{conduct unbecoming a lawyer} (\small freq=1)
  \item [\checkmark] \texttt{double jeopardy} (\small freq=1)
  \item [\checkmark] \texttt{asset forfeiture} (\small freq=1)
  \item [\checkmark] \texttt{negligence element} (\small freq=1)
  \item [\checkmark] \texttt{cretion} (\small freq=1)
  \item [\checkmark] \texttt{2. torts negligence} (\small freq=1)
  \item [\checkmark] \texttt{2. justiciability} (\small freq=1)
  \item [\checkmark] \texttt{cybersecurity law} (\small freq=1)
  \item [\checkmark] \texttt{2. agency theory} (\small freq=1)
  \item[$\times$] \texttt{2. actus reu} (\small freq=1)
  \item [\checkmark] \texttt{assignee} (\small freq=1)
  \item [\checkmark] \texttt{successor liability} (\small freq=1)
  \item[$\times$] \texttt{mysqli connect errno} (\small freq=1)
  \item[$\times$] \texttt{2. ** rules of civil procedure**} (\small freq=1)
  \item [\checkmark] \texttt{2. intent} (\small freq=1)
  \item [\checkmark] \texttt{1. **hearsay rule**} (\small freq=1)
  \item [\checkmark] \texttt{**case law**} (\small freq=1)
  \item [\checkmark] \texttt{2. actual malice} (\small freq=1)
  \item [\checkmark] \texttt{gifts law} (\small freq=1)
  \item [\checkmark] \texttt{crime scene evidence} (\small freq=1)
  \item[$\times$] \texttt{banque maladroit} (\small freq=1)
  \item [\checkmark] \texttt{sequence of preference} (\small freq=1)
  \item [\checkmark] \texttt{administrative law standard of review} (\small freq=1)
  \item [\checkmark] \texttt{2. unconstitutional} (\small freq=1)
  \item [\checkmark] \texttt{model rule 1.6} (\small freq=1)
  \item [\checkmark] \texttt{administrative common law} (\small freq=1)
  \item[$\times$] \texttt{criminal procedure miranda warning} (\small freq=1)
  \item [\checkmark] \texttt{2. in pari delicto} (\small freq=1)
  \item [\checkmark] \texttt{2. **attachment of liens**} (\small freq=1)
  \item[$\times$] \texttt{1. mock trial} (\small freq=1)
  \item[$\times$] \texttt{1. doctrine of agency} (\small freq=1)
  \item [\checkmark] \texttt{**abuse of process**} (\small freq=1)
  \item[$\times$] \texttt{estudiantesformation} (\small freq=1)
  \item[$\times$] \texttt{\textbackslash{}\# administration of estate} (\small freq=1)
  \item [\checkmark] \texttt{post trial relief} (\small freq=1)
  \item[$\times$] \texttt{1. auspicious opportunity} (\small freq=1)
  \item [\checkmark] \texttt{performance} (\small freq=1)
  \item [\checkmark] \texttt{**classification of crimes**} (\small freq=1)
  \item [\checkmark] \texttt{codification} (\small freq=1)
  \item [\checkmark] \texttt{extrinsic fraud} (\small freq=1)
  \item [\checkmark] \texttt{1. qualified immunity} (\small freq=1)
  \item [\checkmark] \texttt{1. choice of law} (\small freq=1)
  \item [\checkmark] \texttt{divorce jurisdiction} (\small freq=1)
  \item [\checkmark] \texttt{**compelled self incrimination**} (\small freq=1)
  \item [\checkmark] \texttt{ucc filing system} (\small freq=1)
  \item [\checkmark] \texttt{animal statute} (\small freq=1)
  \item [\checkmark] \texttt{2. void} (\small freq=1)
  \item[$\times$] \texttt{2. testamentary freedom} (\small freq=1)
  \item [\checkmark] \texttt{1. common law} (\small freq=1)
  \item [\checkmark] \texttt{jurisdictional standing} (\small freq=1)
  \item[$\times$] \texttt{abrogation of rules doctrine} (\small freq=1)
\end{itemize}
\end{multicols}

\paragraph{Low Frequency Concepts (Bottom 25\%, n=50):}

\begin{multicols}{2}
\begin{itemize}[leftmargin=*, itemsep=0pt, parsep=2pt]
  \item [\checkmark] \texttt{seventh amendment} (\small freq=1)
  \item [\checkmark] \texttt{adoption} (\small freq=1)
  \item [\checkmark] \texttt{demurrer} (\small freq=1)
  \item[$\times$] \texttt{mirrors and manoogian} (\small freq=1)
  \item [\checkmark] \texttt{constitutional search and seizure} (\small freq=1)
  \item [\checkmark] \texttt{1. **mistrial**} (\small freq=1)
  \item [\checkmark] \texttt{act childhood} (\small freq=1)
  \item [\checkmark] \texttt{civil liberty} (\small freq=1)
  \item [\checkmark] \texttt{testimony} (\small freq=1)
  \item[$\times$] \texttt{sure, here's a random start} (\small freq=1)
  \item[$\times$] \texttt{1. delegated legislation} (\small freq=1)
  \item[$\times$] \texttt{elle v jaffe} (\small freq=1)
  \item [\checkmark] \texttt{mortgage} (\small freq=1)
  \item[$\times$] \texttt{constitutional adoption of child} (\small freq=1)
  \item [\checkmark] \texttt{judicial interpretation} (\small freq=1)
  \item[$\times$] \texttt{willful blindnes} (\small freq=1)
  \item[$\times$] \texttt{2. cruzan competent} (\small freq=1)
  \item[$\times$] \texttt{admission by exam} (\small freq=1)
  \item [\checkmark] \texttt{miracle} (\small freq=1)
  \item[$\times$] \texttt{commingling of asset} (\small freq=1)
  \item [\checkmark] \texttt{ultra vire} (\small freq=1)
  \item [\checkmark] \texttt{title guarantee} (\small freq=1)
  \item [\checkmark] \texttt{administrative agency law} (\small freq=1)
  \item[$\times$] \texttt{2. mcaptured} (\small freq=1)
  \item[$\times$] \texttt{avoidance of absurdity doctrine} (\small freq=1)
  \item [\checkmark] \texttt{special appearance} (\small freq=1)
  \item [\checkmark] \texttt{environmental law} (\small freq=1)
  \item [\checkmark] \texttt{property law doctrine} (\small freq=1)
  \item [\checkmark] \texttt{legal pleading} (\small freq=1)
  \item [\checkmark] \texttt{u.s. constitution} (\small freq=1)
  \item[$\times$] \texttt{sum and substance rule} (\small freq=1)
  \item [\checkmark] \texttt{willful violation} (\small freq=1)
  \item[$\times$] \texttt{produits right} (\small freq=1)
  \item[$\times$] \texttt{escolar.py} (\small freq=1)
  \item [\checkmark] \texttt{biological mother} (\small freq=1)
  \item [\checkmark] \texttt{trial court jurisdiction} (\small freq=1)
  \item [\checkmark] \texttt{settlement agreement} (\small freq=1)
  \item [\checkmark] \texttt{affinity group} (\small freq=1)
  \item[$\times$] \texttt{2.hair setting} (\small freq=1)
  \item [\checkmark] \texttt{1.naeus} (\small freq=1)
  \item[$\times$] \texttt{2. attorneys' ethic} (\small freq=1)
  \item [\checkmark] \texttt{1. **gambling**} (\small freq=1)
  \item [\checkmark] \texttt{entitlement to counsel} (\small freq=1)
  \item [\checkmark] \texttt{unconstitutional search} (\small freq=1)
  \item [\checkmark] \texttt{fruition of the contract} (\small freq=1)
  \item[$\times$] \texttt{1. un administrable estate} (\small freq=1)
  \item[$\times$] \texttt{astype immunity} (\small freq=1)
  \item [\checkmark] \texttt{ripeness doctrine} (\small freq=1)
  \item[$\times$] \texttt{enable refreshment license} (\small freq=1)
  \item [\checkmark] \texttt{2. de facto merger} (\small freq=1)
\end{itemize}
\end{multicols}

\end{document}